\documentclass[runningheads]{llncs}
\usepackage{graphicx}

\usepackage{tikz}
\usepackage{comment} 
\usepackage{amsmath,amssymb} 
\usepackage{color}

\usepackage{subfig}
\usepackage{array}
\usepackage{multirow}

\usepackage[hidelinks,colorlinks=true]{hyperref}
\usepackage{pifont}
\newcommand{\cmark}{\ding{51}}
\newcommand{\xmark}{\ding{55}}

\newcommand{\etal}{\emph{et al. }}
\newcommand{\eg}{\emph{e.g., }}
\newcommand{\ie}{\emph{i.e., }}

\begin{document}
\pagestyle{headings}
\mainmatter
\def\ECCVSubNumber{3400}  

\title{LiteFlowNet3: Resolving Correspondence Ambiguity for More Accurate Optical Flow Estimation} 

\titlerunning{Resolving Correspondence Ambiguity for More Accurate Flow Estimation}
%
\author{Tak-Wai Hui\inst{1}\orcidID{0000-0002-1441-9289} \and
Chen Change Loy\inst{2}\orcidID{0000-0001-5345-1591}}
\authorrunning{T.-W. Hui \and C. C. Loy}
%
\institute{{The Chinese University of Hong Kong \and
Nanyang Technological University} \\
\url{https://github.com/twhui/LiteFlowNet3} \\
\email{twhui@ie.cuhk.edu.hk, ccloy@ntu.edu.sg}}
\maketitle

\begin{abstract}
Deep learning approaches have achieved great success in addressing the problem of optical flow estimation. 
The keys to success lie in the use of cost volume and coarse-to-fine flow inference. 
However, the matching problem becomes ill-posed when partially occluded or homogeneous regions exist in images. 
This causes a cost volume to contain outliers and affects the flow decoding from it. 
Besides, the coarse-to-fine flow inference demands an accurate flow initialization. Ambiguous correspondence yields erroneous flow fields and affects the flow inferences in subsequent levels.
In this paper, we introduce LiteFlowNet3, a deep network consisting of two specialized modules, to address the above challenges.
(1) We ameliorate the issue of outliers in the cost volume by amending each cost vector through an adaptive modulation prior to the flow decoding. 
(2) We further improve the flow accuracy by exploring local flow consistency. To this end, each inaccurate optical flow is replaced with an accurate one from a nearby position through a novel warping of the flow field. 
LiteFlowNet3 not only achieves promising results on public benchmarks but also has a small model size and a fast runtime.

\end{abstract}

\section{Introduction}
Optical flow estimation is a classical problem in computer vision. It is widely used in many applications such as motion tracking, action recognition, video segmentation, 3D reconstruction, and more. 
With the advancement of deep learning, many research works have attempted to address the problem by using convolutional neural networks (CNNs)~\cite{Hui18,Hui20,Hur19,Ilg17,Liu19,Sun18,Sun19,Yang19,Yin19}. The majority of the CNNs belongs to the 2-frame method that infers a flow field from an image pair. 
Particularly, LiteFlowNet~\cite{Hui18} and PWC-Net~\cite{Sun18} are the first CNNs to propose using the feature warping and cost volume at multiple pyramid levels in a coarse-to-fine estimation. This greatly reduces the number of model parameters from 160M in FlowNet2~\cite{Ilg17} to 5.37M in LiteFlowNet and 8.75M in PWC-Net while accurate flow estimation is still maintained.

One of the keys to success for the lightweight optical flow CNNs is the use of cost volume for establishing correspondence at each pyramid level.
However, a cost volume is easily corrupted by ambiguous feature matching~\cite{Kang01,Rhemann11,Xu17}. This causes flow fields that are decoded from the cost volume to become unreliable.
The underlying reasons for the existence of ambiguous matching are twofold.
First, when given a pair of images, it is impossible for a feature point in the first image to find the corresponding point in the second image, when the latter is occluded. 
Second, ambiguous correspondence is inevitable in homogeneous regions (\eg shadows, sky, and walls) of images. 
Another key to success for the optical flow CNNs is to infer flow fields using a coarse-to-fine framework. However, this approach highly demands an accurate flow initialization from the preceding pyramid level. Once ambiguous correspondence exists, erroneous optical flow is generated and propagates to subsequent levels.

To address the aforementioned challenges, we attempt to make correspondence across images less ambiguous and in turn improves the accuracy of optical flow CNNs by introducing the following specialized CNN modules:

\noindent 
\textbf{Cost Volume Modulation.} Ambiguous feature matching causes outliers to exist in a cost volume. Inaccurate cost vectors need to be amended to allow the correct flow decoding.
To deal with occlusions, earlier work improves the matching process by using the offset-centered matching windows~\cite{Kang01}. A cost volume is filtered to remove outliers prior to the correspondence decoding~\cite{Rhemann11,Xu17}. 
However, existing optical flow CNNs~\cite{Hui20,Hur19,Ilg17,Liu19,Sun19,Yin19,Yang19} infer optical flow from a cost volume using convolutions without explicitly addressing the issue of outliers. 
We propose to amend each cost vector in the cost volume by using an adaptive affine transformation. A confidence map that pinpoints the locations of unreliable flow is used to facilitate the generation of transformation parameters. 

\noindent\textbf{Flow Field Deformation.} When the correspondence problem becomes ill-posed, it is very difficult to find correct matching pairs. Local flow consistency and co-occurrence between flow boundaries and intensity edges are commonly used as the clues to regularize flow fields in conventional methods~\cite{Werlberger09,Zimmer11}. The two principles are also adopted in recent optical flow CNNs~\cite{Hui18,Hui20,Hur19}.
We propose a novel technique to further improve the flow accuracy by using the clue from local flow consistency. Intuitively, we replace each inaccurate optical flow with an accurate one from a nearby position having similar feature vectors. The replacement is achieved by a meta-warping of the flow field in accordance with a computed displacement field (similar to optical flow but the displacement field no longer represents correspondence). We compute the displacement field by using a confidence-guided decoding from an auto-correlation cost volume. 

In this work, we make the first attempt to use cost volume modulation and flow field deformation in optical flow CNNs.
We extend our previous work (LiteFlowNet2~\cite{Hui20}) by incorporating the proposed modules for addressing the aforementioned challenges. 
LiteFlowNet3 achieves promising performance in the 2-frame method. It outperforms VCN-small~\cite{Yang19}, IRR-PWC~\cite{Hur19}, PWC-Net+~\cite{Sun19}, and LiteFlowNet2 on Sintel and KITTI.
Even though SelFlow~\cite{Liu19} (a multi-frame method) and HD$^{3}$~\cite{Yin19} use extra training data, LiteFlowNet3 outperforms SelFlow on Sintel clean and KITTI while it performs better than HD$^{3}$ on Sintel, KITTI 2012, and KITTI 2015 (in foreground region). 
LiteFlowNet3 does not suffer from the artifact problem on real-world images as HD$^{3}$, while being 7.7 times smaller in model size and 2.2 times faster in runtime. 

\section{Related Work}
\noindent\textbf{Variational Approach.} Since the pioneering work of Horn and Schunck~\cite{Horn81}, the variational approach has been widely studied for optical flow estimation. Brox \etal address the problem of illumination change across images by introducing the gradient constancy assumption~\cite{Brox04}. Brox \etal \cite{Brox04} and Papenberg \etal \cite{Papenberg06} propose the use of image warping in minimizing an energy functional. Bailer \etal propose Flow Fields~\cite{Bailer15}, which is a searching-based method. Optical flow is computed by a numerical optimization with multiple propagations and random searches. In EpicFlow~\cite{Revaud15}, Revaud \etal use sparse flows as an initialization and then interpolate them to a dense flow field by fitting a local affine model at each pixel based on nearby matches. The affine parameters are computed as the least-square solution of an over-determined system. Unlike EpicFlow, we use an adaptive affine transformation to amend a cost volume. The transformation parameters are implicitly generated in the CNN instead.

\noindent\textbf{Cost Volume Approach.} Kang \etal address the problem of ambiguous matching by using the offset-centered windows and select a subset of neighboring image frames to perform matching dynamically~\cite{Kang01}.
Rhemann \etal propose to filter a cost volume using an edge-preserving filter~\cite{Rhemann11}. 
In DCFlow~\cite{Xu17}, Xu \etal exploit regularity in a cost volume and improve the optical flow accuracy by adapting the semi-global matching. 
With the inspiration of improving cost volume from the above conventional methods, we propose to modulate each cost vector in the cost volume by using an affine transformation prior to the flow decoding. The transformation parameters are adaptively constructed to suit different cost vectors. 
In particular, DCFlow combines the interpolation in EpicFlow~\cite{Revaud15} with a complementary scheme to convert a sparse correspondence to a dense one. On the contrary, LiteFlowNet3 applies an affine transformation to all elements in the cost volume but not to sparse correspondence. 

\noindent \textbf{Unsupervised and Self-supervised Optical Flow Estimation.} To avoid annotating labels, Meister \etal propose a framework that uses the difference between synthesized and real images for unsupervised training~\cite{Meister18}. 
Liu \etal propose SelFlow that distills reliable flow estimations from non-occluded pixels in a large dataset using self-supervised training~\cite{Liu19}. It also uses multiple frames and fine-tunes the self-supervised model in supervised training for improving the flow accuracy further. 
Unlike the above works, we focus on supervised learning. Even though LiteFlowNet3 is a 2-frame method and   trained on a much smaller dataset, it still outperforms SelFlow on Sintel clean and KITTI. 

\noindent \textbf{Supervised Learning of Optical Flow.} Dosovitskiy \etal develop FlowNet~\cite{Dosovitskiy15}, the first optical flow CNN. 
Mayer \etal extend FlowNet to estimate disparity and scene flow~\cite{Mayer16}.
In FlowNet2~\cite{Ilg17}, Ilg \etal improve the flow accuracy of FlowNet by cascading several variants of it. However, the model size is increased to over 160M parameters and it also demands a high computation time. 
Ranjan \etal develop a compact network SPyNet~\cite{Ranjan17}, but the accuracy is not comparable to FlowNet2. 
Our LiteFlowNet~\cite{Hui18}, which consists of the cascaded flow inference and flow regularization, has a small model size (5.37M) and comparable performance as FlowNet2. We then develop LiteFlowNet2 for more accurate flow accuracy and faster runtime~\cite{Hui20}. LiteFlowNet3 is built upon LiteFlowNet2 with the incorporation of cost volume modulation and flow field deformation for improving the flow accuracy further.
A concurrent work to LiteFlowNet is PWC-Net~\cite{Sun18}, which proposes using the feature warping and cost volume as LiteFlowNet. Sun \etal then develop PWC-Net+ by improving the training protocol~\cite{Sun19}. 
Ilg \etal extend FlowNet2 to FlowNet3 with the joint learning of occlusion and optical flow~\cite{Ilg18}. 
In Devon~\cite{Lu20}, Lu \etal perform feature matching that is governed by an external flow field. On the contrary, our displacement field is used to deform optical flow but not to facilitate feature matching.
Hur \etal propose IRR-PWC~\cite{Hur19}, which improves PWC-Net by adopting the flow regularization from LiteFlowNet as well as introducing the occlusion decoder and weight sharing. 
Yin \etal introduce HD$^{3}$ for learning a probabilistic pixel correspondence~\cite{Yin19}, but it requires pre-training on ImageNet. While LiteFlowNet3 learns a flow confidence implicitly but not computed from the probabilistic estimation. Despite HD$^{3}$ uses extra training data and 7.7 times more parameters, LiteFlowNet3 outperforms HD$^{3}$ on Sintel, KITTI 2012, and KITTI 2015 (in foreground region).
LiteFlowNet3 outperforms VCN-small~\cite{Yang19} even though the model sizes are similar.
Comparing to deformable convolution~\cite{Dai17}, we perform deformation on flow fields but not on feature maps. Our deformation aims to replace each inaccurate optical flow with an accurate one from a nearby position in the flow field, while deformable convolution aims to augment spatial sampling.

\section{LiteFlowNet3}
\begin{figure*}[t]
\centering
   \includegraphics[width=\linewidth]{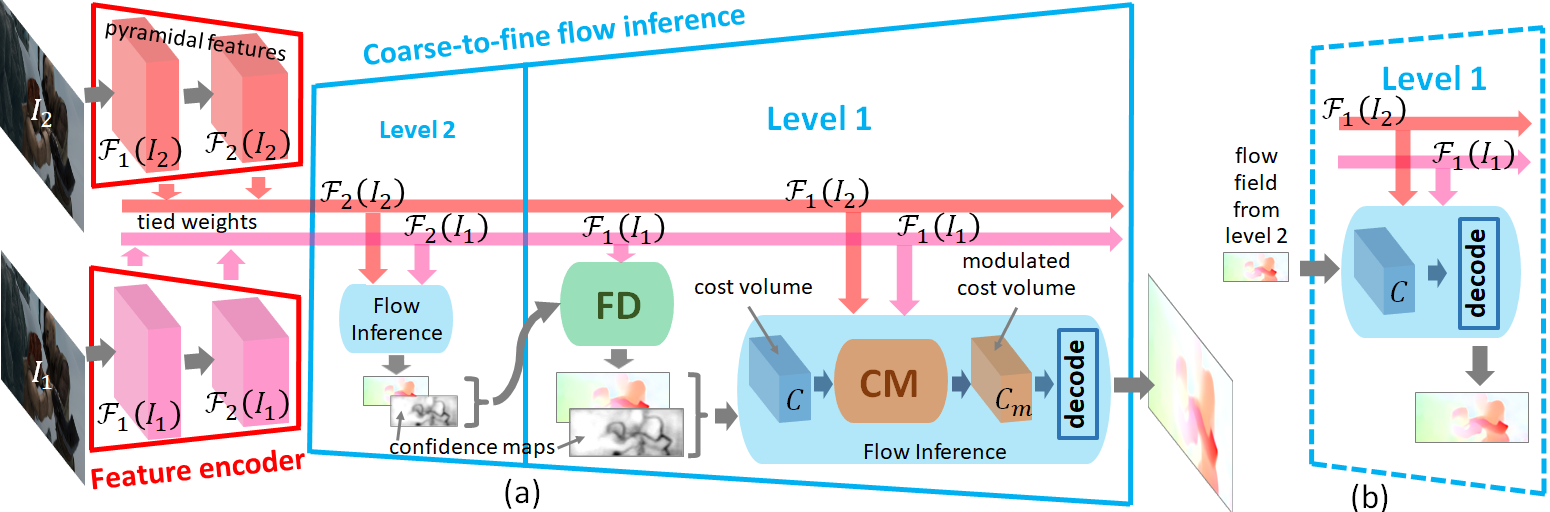}
\caption{(a) A simplified overview of LiteFlowNet3. Flow field deformation (FD) and cost volume modulation (CM) together with confidence maps are incorporated into LiteFlowNet3. For the ease of presentation, only a 2-level encoder-decoder structure is shown. The proposed modules are applicable to other levels but not limited to level 1. (b) The optical flow inference in LiteFlowNet2~\cite{Hui20}.}
\label{fig:liteflownet3}
\end{figure*}

Feature matching becomes ill-posed in homogeneous and partially occluded regions as one-to-multiple correspondence occurs for the first case while one-to-none correspondence occurs for the second case. Duplicate of image structure (so-called ``ghosting effect'') is inevitable whenever warping is applied to images~\cite{Janai18}. The same also applies to feature maps. In coarse-to-fine estimation, erroneous optical flow resulting from the preceding level affects the subsequent flow inferences.
To address the above challenges, we develop two specialized CNN modules: \textit{Cost volume Modulation} (CM) and \textit{Flow field Deformation} (FD). We demonstrate the applicability of the modules on LiteFlowNet2~\cite{Hui20}. The resulting network is named as LiteFlowNet3. Figure~\ref{fig:liteflownet3} illustrates a simplified overview of the network architecture.
FD is used to refine the previous flow estimate before it is used as a flow initialization in the current pyramid level.
In flow inference, the cost volume is amended by CM prior to the flow decoding. 

\subsection{Preliminaries}
We first provide a concise description on the construction of cost volume in optical flow CNNs. Suppose a pair of images $I_{1}$ (at time $t = 1$) and $I_{2}$ (at time $t = 2$) is given. We convert $I_{1}$ and $I_{2}$ respectively into pyramidal feature maps ${\mathcal F}_{1}$ and ${\mathcal F}_{2}$ through a feature encoder. We denote ${\bf x}$ as a point in the rectangular domain~$\Omega \subset~\mathbb{R}^{2}$.  
Correspondence between $I_{1}$ and $I_{2}$ is established by computing the dot product between two high-level feature vectors in the individual feature maps ${\mathcal F}_{1}$ and ${\mathcal F}_{2}$ as follows~\cite{Dosovitskiy15}:
\begin{equation}
c({\bf x};D) = {\mathcal F}_{1}({\bf x}) \cdot {\mathcal F}_{2}({\bf x}')/N,
\label{eq:correlation}
\end{equation}
where $D$ is the maximum matching radius, $c({\bf x};D)$ (a 3D column vector with length $2D+1$) is the collection of matching costs between feature vectors ${\mathcal F}_{1}({\bf x})$ and ${\mathcal F}_{2}({\bf x}')$ for all possible ${\bf x}'$ such that $\|{\bf x} - {\bf x}'\|_{\infty} = D$, and $N$ is the length of the feature vector. 
Cost volume $C$ is constructed by aggregating all $c({\bf x};D)$ into a 3D grid. Flow decoding is then performed on $C$ using convolutions (or native winner-takes-all approach~\cite{Kang01}). The resulting flow field ${\bf u}: \Omega \rightarrow \mathbb{R}^{2}$ provides the dense correspondence from $I_{1}$ to $I_{2}$.
In the following, we will omit variable $D$ that indicates the maximum matching radius for brevity and use $c({\bf x})$ to represent the cost vector at {\bf x}. When we discuss operations in a pyramid level, the same operations are applicable to other levels.

\subsection{Cost Volume Modulation}
\label{sec:cost volume modulation}
\begin{figure*}[t]
\centering
	\includegraphics[height=2.5cm]{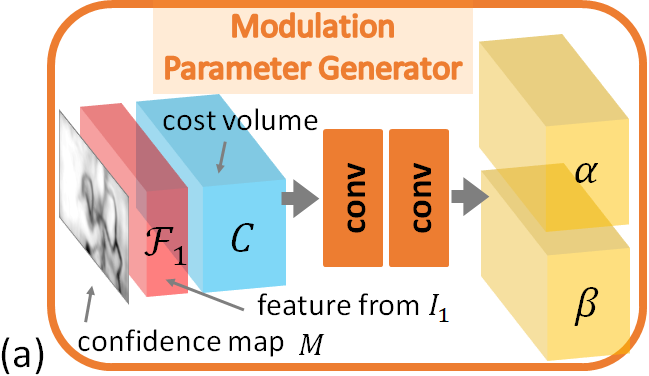}
	\includegraphics[height=2.5cm]{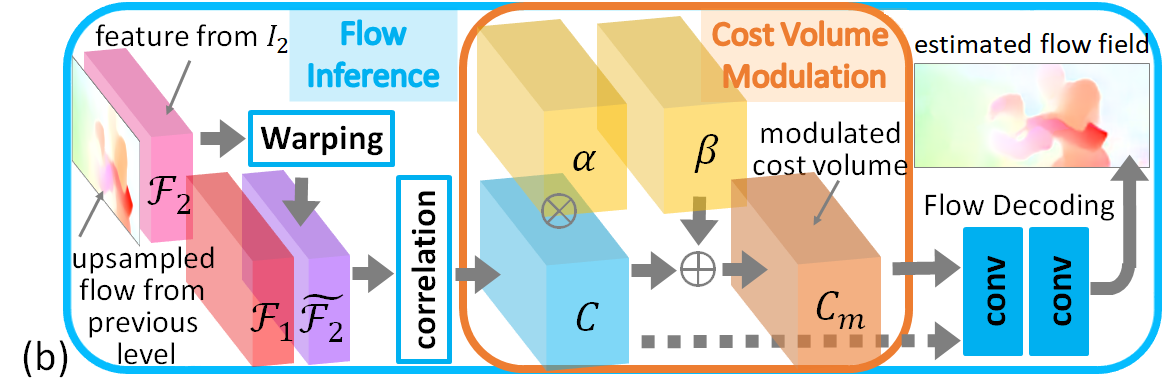} 
\caption{(a) Modulation tensors ($\alpha$, $\beta$) are adaptively constructed for each cost volume. (b) Cost volume modulation is integrated into the flow inference. Instead of leaving cost volume $C$ unaltered (via the dashed arrow), it is amended to $C_{m}$ by using the adaptive modulation prior to the flow decoding. Note: ``conv" denotes several convolution layers.}
\label{fig:cost volume modulation}
\end{figure*}

Given a pair of images, the existence of partial occlusion and homogeneous regions makes the establishment of correspondence very challenging. This situation also occurs on feature space because simply transforming images into feature maps does not resolve the correspondence ambiguity. 
In this way, a cost volume is corrupted and the subsequent flow decoding is seriously affected. 
Conventional methods~\cite{Rhemann11,Xu17} address the above problem by filtering a cost volume prior to the decoding. 
But there has not been any existing works to address this problem for optical flow CNNs.
Some studies~\cite{Brabandere16,Hui18,Hur19} have revealed that applying feature-driven convolutions on feature space is an effective approach to influence the feed-forward behavior of a network since the filter weights are adaptively constructed. Therefore, we devise to filter outliers in a cost volume by using an adaptive modulation. We will show that our modulation approach is not only effective in improving the flow accuracy but also parameter-efficient.

An overview of cost volume modulation is illustrated in Fig.~\ref{fig:cost volume modulation}\textcolor{red}{b}. At a pyramid level, each cost vector $c({\bf x})$ in cost volume $C$ is adaptively modulated by an affine transformation ($\alpha({\bf x}), \beta({\bf x})$) as follows:
\begin{equation}
c_{m}({\bf x}) = \alpha({\bf x}) \otimes c({\bf x}) \oplus \beta({\bf x}),
\end{equation}
where $c_{m}({\bf x})$ is the modulated cost vector, ``$\otimes$'' and ``$\oplus$'' denote element-wise multiplication and addition, respectively.
The dimension of the modulated cost volume is same as the original. This property allows cost volume modulation to be jointly used and trained with an existing network without major changes made to the original network architecture.

To have an efficient computation, the affine parameters $\{\alpha({\bf x}), \beta({\bf x})\}, \forall {\bf x} \in \Omega$, are generated altogether in the form of modulation tensors ($\alpha, \beta$) having the same dimension as $C$. As shown in Fig.~\ref{fig:cost volume modulation}\textcolor{red}{a}, we use cost volume $C$, feature ${\mathcal F}_{1}$ from the encoder, and confidence map $M$ at the same pyramid level as the inputs to the modulation parameter generator.
The confidence map is introduced to facilitate the generation of modulation parameters. Specifically, $M({\bf x})$ pinpoints the probability of having an accurate optical flow at ${\bf x}$ in the associated flow field. The confidence map is constructed by introducing an additional output in the preceding optical flow decoder. A sigmoid function is used to constraint its values to $[0, 1]$. We train the confidence map using a L2 loss with the ground-truth label $M_{gt}({\bf x})$ as follows:
\begin{equation}
M_{gt}({\bf x}) = \text{e}^{-\|{\bf u}_{gt}({\bf x}) - {\bf u}({\bf x})\|^{2}},
\end{equation}
where ${\bf u}_{gt}({\bf x})$ is the ground truth of ${\bf u}({\bf x})$. An example of predicted confidence maps will be provided in Section~\ref{sec:ablation study}. 

\begin{figure}[t]
\centering
\captionsetup[subfigure]{labelformat=empty}
  \subfloat[(a) NO: LiteFlowNet2~\cite{Hui20}\label{fig:NO}]{\includegraphics[height=4.0cm]{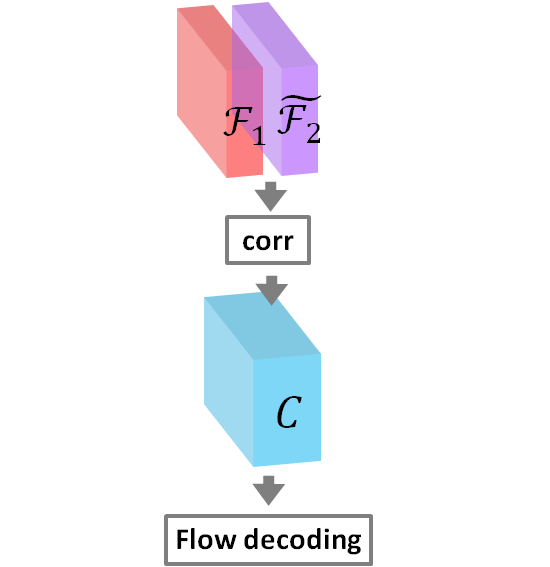}}
  \subfloat[(b) FF: PWC-Net+~\cite{Sun19}\label{fig:FF}]{\includegraphics[height=4.0cm]{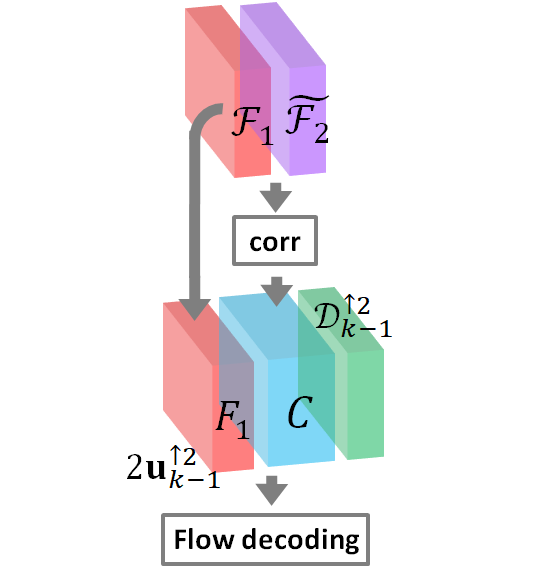}}
   \subfloat[(c) Ours: LiteFlowNet3\label{fig:Ours}]{\includegraphics[height=4.0cm]{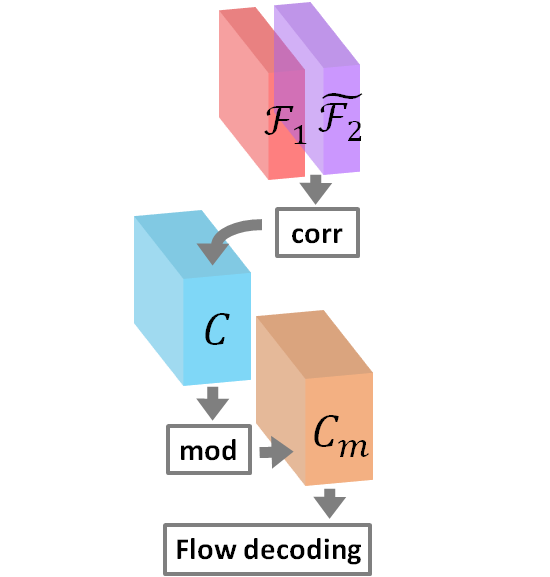}}
\caption{Augmenting a cost volume under different configurations. (a) NO, (b) FF, and (c) Our solution: Cost volume $C$ is modulated to $C_{m}$ by using an adaptive affine transformation prior to the flow decoding. Note: ``corr" and ``mod" denote correlation and modulation, respectively. Correlation is performed on $\mathcal{F}_{1}$ and warped $\mathcal{F}_{2}$ (\ie $\widetilde{\mathcal{F}_{2}}$).}
\label{fig:flow decoding}
\end{figure}

\begin{table}[t]
\centering
\caption{Average end-point error (AEE) and model size of different models trained on FlyingChairs under different augmentations of cost volume.} \label{tab:results of different flow decodings}
\begin{tabular}{|cccc|}
\hline
\multicolumn{1}{|l|}{Augmentations}
&\multicolumn{1}{c|}{NO}
&\multicolumn{1}{c|}{FF}
&\multicolumn{1}{c|}{Ours}\\
\hline
\multicolumn{1}{|l|}{Features of $I_{1}$}
&\multicolumn{1}{c|}{\xmark}
&\multicolumn{1}{c|}{\cmark}
&\multicolumn{1}{c|}{\xmark}\\

\multicolumn{1}{|l|}{Flow field}
&\multicolumn{1}{c|}{\xmark}
&\multicolumn{1}{c|}{\cmark}
&\multicolumn{1}{c|}{\xmark}\\

\multicolumn{1}{|l|}{Modulation}
&\multicolumn{1}{c|}{\xmark}
&\multicolumn{1}{c|}{\xmark}
&\multicolumn{1}{c|}{\cmark}\\
\hline\hline

\multicolumn{1}{|l|}{Number of model parameters (M)}
&\multicolumn{1}{c|}{\textbf{6.42}}
&\multicolumn{1}{c|}{7.16}
&\multicolumn{1}{c|}{7.18}\\

\multicolumn{1}{|l|}{Sintel Clean (training set)}
&\multicolumn{1}{c|}{2.71}
&\multicolumn{1}{c|}{2.70}
&\multicolumn{1}{c|}{\textbf{2.65}}\\

\multicolumn{1}{|l|}{Sintel Final (training set)}
&\multicolumn{1}{c|}{4.14}
&\multicolumn{1}{c|}{4.20}
&\multicolumn{1}{c|}{\textbf{4.02}}\\

\multicolumn{1}{|l|}{KITTI 2012 (training set)}
&\multicolumn{1}{c|}{4.20}
&\multicolumn{1}{c|}{4.28}
&\multicolumn{1}{c|}{\textbf{3.95}}\\

\multicolumn{1}{|l|}{KITTI 2015 (training set)}
&\multicolumn{1}{c|}{11.12}
&\multicolumn{1}{c|}{11.30}
&\multicolumn{1}{c|}{\textbf{10.65}}\\
\hline 
\end{tabular}
\end{table}

\vspace{0.1cm} \noindent \textbf{Discussion.} In the literature, there are two major approaches to infer a flow field from a cost volume as shown in Fig.~\ref{fig:flow decoding}.
The first approach (Fig.~\ref{fig:NO}) is to perform flow decoding directly on the cost volume without any augmentation~\cite{Hui18,Hui20}. This is similar to the conventional winner-takes-all approach~\cite{Kang01} except using convolutions for yielding flow fields rather than argument of the minimum.
The second approach (Fig.~\ref{fig:FF}) feed-forwards the pyramidal features $\mathcal{F}_{1}$ from the feature encoder~\cite{Sun18,Sun19}. It also feed-forwards the upsampled flow field ($2{\bf u}_{k-1}^{\uparrow 2}$) and features (${\mathcal D}_{k-1}^{\uparrow 2}$) from the previous flow decoder (at level $k-1$). Flow decoding is then performed on the concatenation. 
Our approach (Fig.~\ref{fig:Ours}) is to perform modulation on the cost volume prior to the flow decoding.
The effectiveness of the above approaches has not been studied in the literature. Here, we use LiteFlowNet2~\cite{Hui20} as the backbone architecture and train all the models from scratch on FlyingChairs dataset~\cite{Dosovitskiy15}. Table~\ref{tab:results of different flow decodings} summarizes the results of our evaluation. Even though FF needs 11.5\% more model parameters than NO, it attains lower flow accuracy. On the contrary, our modulation approach that has just 0.28\% more parameters than FF outperforms the compared methods on all the benchmarks, especially KITTI 2012 and KITTI 2015. This indicates that a large CNN model does not always perform better than a smaller one.

\begin{figure}[t]
\centering
   \includegraphics[height=2.6cm]{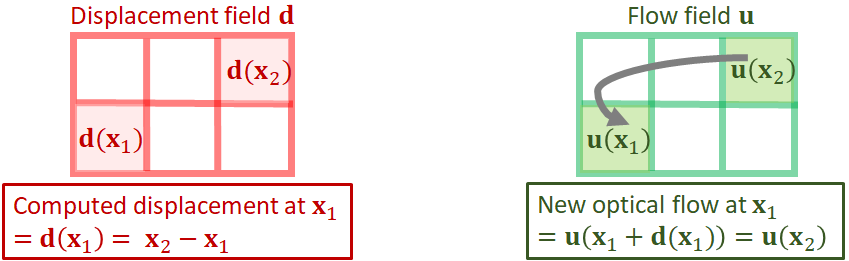}
\caption{Replacing an inaccurate optical flow ${\bf u}({\bf x}_{1})$ with an accurate optical flow ${\bf u}({\bf x}_{2})$ through a meta-warping governed by displacement ${\bf d}({\bf x}_{1})$.}
\label{fig:flow field deformation principle}
\end{figure}

\subsection{Flow Field Deformation}
\label{sec:flow field deformation}
In coarse-to-fine flow estimation, a flow estimate from the preceding decoder is used as a flow initialization for the subsequent decoder. This highly demands the previous estimate to be accurate. Otherwise, erroneous optical flow is propagated to subsequent levels and affects the flow inference. 
Using cost volume modulation alone is not able to address this problem. We explore local flow consistency~\cite{Werlberger09,Zimmer11} and propose to use a meta-warping for improving the flow accuracy. 

Intuitively, we refine a given flow field by replacing each inaccurate optical flow with an accurate one from a nearby position using the principle of local flow consistency. As shown in Fig.~\ref{fig:flow field deformation principle}, suppose an optical flow ${\bf u}({\bf x}_{1})$ is inaccurate. With some prior knowledge, 1) a nearby optical flow ${\bf u}({\bf x}_{2})$ such that ${\bf x}_{2} = {\bf x}_{1} + {\bf d}({\bf x}_{1})$ is known to be accurate as indicated by a confidence map; 2) the pyramidal features of $I_{1}$ at ${\bf x}_{1}$ and ${\bf x}_{2}$ are similar \ie $\mathcal{F}_{1}({\bf x}_{1}) \sim \mathcal{F}_{1}({\bf x}_{2})$ as indicated by an auto-correlation cost volume. Since image points that have similar feature vectors have similar optical flow in a neighborhood, we replace ${\bf u}({\bf x}_{1})$ with a clone of ${\bf u}({\bf x}_{2})$. 

The previous analysis is just for a single flow vector. To cover the whole flow field, we need to find a displacement vector for every position in the flow field. In other words, we need to have a displacement field for guiding the meta-warping of flow field. We use a warping mechanism that is similar to image~\cite{Ilg17} and feature warpings~\cite{Hui18,Sun18}. The differences are that our meta-warping is limited to two channels and the physical meaning of the introduced displacement field no longer represents correspondence across images.

\begin{figure*}[t]
\centering
   \includegraphics[height=2.2cm]{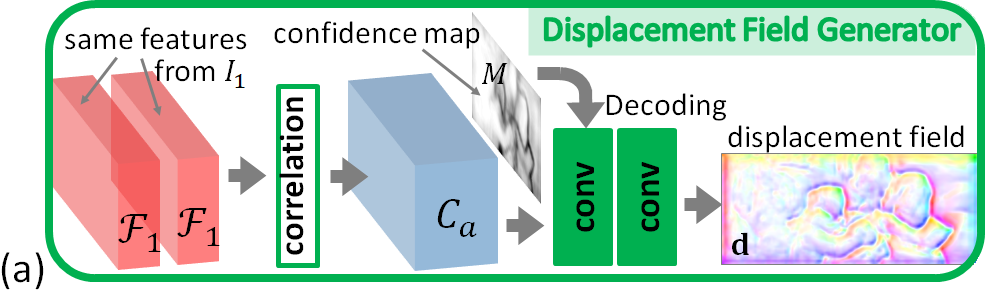}
   \includegraphics[height=2.2cm]{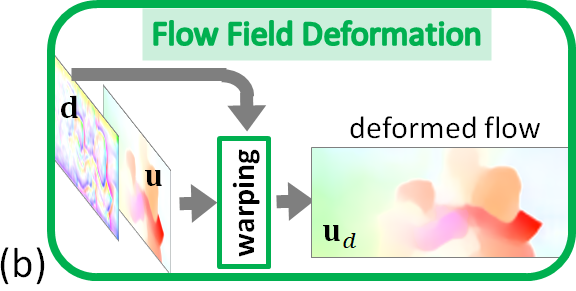}
\caption{(a) Displacement field ${\bf d}$ is constructed according to auto-correlation cost volume $C_{a}$ and confidence map $M$. (b) Flow field ${\bf u}$ is warped to ${\bf u}_{d}$ in accordance to ${\bf d}$. Flow deformation is performed before ${\bf u}$ is used as an initialization for the flow inference. Note: ``conv" denotes several convolution layers.}
\label{fig:flow field deformation}
\end{figure*}

An overview of flow field deformation is illustrated in Fig.~\ref{fig:flow field deformation}\textcolor{red}{b}. At a pyramid level, we replace ${\bf u}({\bf x})$ with an neighboring optical flow by warping of ${\bf u}({\bf x})$ in accordance to the computed displacement ${\bf d}({\bf x})$ as follows:
\begin{equation}
{\bf u}_{d}({\bf x}) = {\bf u}\left({\bf x} + {\bf d}({\bf x})\right).
\end{equation}
In particular, not every optical flow needs an amendment. Suppose ${\bf u}({\bf x_{0}})$ is very accurate, then no flow warping is required \ie ${\bf d}({\bf x_{0}}) \sim {\bf 0}$. 

To generate the displacement field, the location of image point having similar feature as the targeted image point needs to be found. This is accomplished by decoding from an auto-correlation cost volume. The procedure is similar to flow decoding from a normal cost volume~\cite{Dosovitskiy15}. As shown in Fig.~\ref{fig:flow field deformation}\textcolor{red}{a}, we first measure the feature similarity of targeted point at ${\bf x}$ and its surrounding points at ${\bf x}'$ by computing auto-correlation cost vector $c_{a}({\bf x}; D)$ between features $\mathcal F_{1}({\bf x})$ and $\mathcal F_{1}({\bf x}')$ as follows:
\begin{equation}
c_{a}({\bf x};D) = {\mathcal F}_{1}({\bf x}) \cdot {\mathcal F}_{1}({\bf x}')/N,
\end{equation}
where $D$ is the maximum matching radius, ${\bf x}$ and ${\bf x}'$ are constrained by $\|{\bf x} - {\bf x}'\|_{\infty} = D$, and $N$ is the length of the feature vector. The above equation is identical to Eq.~\eqref{eq:correlation} except using features from $I_{1}$ only. Auto-correlation cost volume $C_{a}$ is then built by aggregating all cost vectors into a 3D grid.

To avoid trivial solution, confidence map $M$ associated with flow field ${\bf u}$ that is constructed by the preceding flow decoder (same as the one presented in Section~\ref{sec:cost volume modulation}) is used to guide the decoding of displacement from $C_{a}$.
As shown in Fig.~\ref{fig:flow field deformation}\textcolor{red}{a}, we use cost volume $C_{a}$ for the auto-correlation of ${\mathcal F}_{1}$ and confidence map $M$ at the same pyramid level as the inputs to the displacement field generator. Rather than flow decoding from the cost volume as the normal descriptor matching~\cite{Dosovitskiy15}, our displacement decoding is performed on the auto-correlation cost volume and is guided by the confidence map. 

\section{Experiments} \label{sec:experiments}
\noindent \textbf{Network Details.} LiteFlowNet3 is built upon LiteFlowNet2~\cite{Hui20}. Flow inference is performed from levels 6 to 3 (and 2) with the given image resolution as level 1. Flow field deformation is applied prior to the cascaded flow inference while cost volume modulation is applied in the descriptor matching unit.
We do not apply the modules to level 6 as no significant improvement on flow accuracy can be observed (and level 2 due to large computational load).
Each module uses four 3$\times$3 convolution layers followed by a leaky rectified linear unit except to use a 5$\times$5 filter in the last layer at levels 4 and 3.
Confidence of flow prediction is implicitly generated by introducing an additional convolution layer in a flow decoder.
Weight sharing is used on the flow decoders and proposed modules. This variant is denoted by the suffix ``S''.

\noindent \textbf{Training Details.} For a fair comparison, we use the same training sets as other optical flow CNNs in the literature~\cite{Dosovitskiy15,Hui18,Hui20,Hur19,Ilg17,Lu20,Ranjan17,Sun18,Sun19,Yang19}. We use the same training protocol (including data augmentation and batch size) as LiteFlowNet2~\cite{Hui20}. We first train LiteFlowNet2 on FlyingChairs dataset~\cite{Dosovitskiy15} using the stage-wise training procedure~\cite{Hui20}. We then integrate brand new modules, cost volume deformation and flow field modulation, into LiteFlowNet2 to form LiteFlowNet3. The newly introduced CNN modules are trained with a learning rate of 1e-4 while the other components are trained with a reduced learning rate of 2e-5 for 300K iterations. We then fine-tune the whole network on FlyingThings3D~\cite{Mayer16} with a learning rate 5e-6 for 500K iterations. Finally, we fine-tune LiteFlowNet3 respectively on a mixture of Sintel~\cite{Butler12} and KITTI~\cite{Menze15}, and KITTI training sets with a learning rate \text{5e-5} for 600K iterations. The two models  are also re-trained with reduced learning rates and iterations same as LiteFlowNet2. 

\subsection{Results}
\label{sec:results}
We evaluate LiteFlowNet3 on the popular optical flow benchmarks including Sintel clean and final passes~\cite{Butler12}, KITTI 2012~\cite{Geiger12}, and KITTI 2015~\cite{Menze15}. We report average end-point error (AEE) for all the benchmarks unless otherwise explicitly specified. More results are available in the supplementary material~\cite{Hui20a}.
 
\noindent\textbf{Preliminary Discussion.} The majority of optical flow CNNs including LiteFlowNet3 are 2-frame methods and use the same datasets for training. However, HD$^{3}$~\cite{Yin19} is pre-trained on ImageNet ($>$ 10M images). SelFlow~\cite{Liu19} uses Sintel movie ($\sim$ 10K images) and multi-view extensions of KITTI ($>$ 20K images) for self-supervised training. SENSE~\cite{Jiang19} uses SceneFlow dataset~\cite{Mayer16} ($>$ 39K images) for pre-training. While SelFlow also uses more than two frames to boost the flow accuracy. Therefore, their evaluations are not directly comparable to the majority of the optical flow CNNs in the literature.

\noindent\textbf{Quantitative Results.} Table~\ref{tab:results} summarizes the AEE results of LiteFlowNet3 and the state-of-the-art methods on the public benchmarks.
With the exception of HD$^{3}$~\cite{Yin19}, SelFlow~\cite{Liu19}, and SENSE~\cite{Jiang19}, all the compared CNN models are trained on the same datasets and are the 2-frame method. Thanks to the cost volume modulation and flow field deformation, LiteFlowNet3 outperforms these CNN models including the recent state-of-the-art methods IRR-PWC~\cite{Hur19} and VCN-small~\cite{Yang19} on both Sintel and KITTI benchmarks.
Despite the recent state-of-the-art methods HD$^{3}$ and SelFlow (a multi-frame method) use extra training data, LiteFlowNet3 outperforms HD$^{3}$ on Sintel, KITTI 2012, and KITTI 2015 (Fl-fg). Our model also performs better than SelFlow on Sintel clean and KITTI. It should be noted that LiteFlowNet3 has a smaller model size and a faster runtime than HD$^{3}$ and VCN~\cite{Yang19} (a larger variant of VCN-small). 
We also perform evaluation by dividing AEE into matched and unmatched regions (error over regions that are visible in adjacent frames or only in one of two adjacent frames, respectively). As revealed in Table~\ref{tab:detailed sintel}, LiteFlowNet3 achieves the best results on both matched and unmatched regions. Particularly, there is a large improvement on unmatched regions comparing to LiteFlowNet2. This indicates that the proposed modules are effective in addressing correspondence ambiguity.

\begin{table*}[t]
\centering
\caption{AEE results on the public benchmarks. (Notes: The values in parentheses are the results of the networks on the data they were trained on, and hence are not directly comparable to the others. The best in each category is in bold and the second best is underlined. For KITTI 2012, ``All'' (or ``Noc'') represents the average end-point error in total (or non-occluded areas). For KITTI 2015, ``Fl-all'' (or ``-fg'') represents the percentage of outliers averaged over all (or foreground) pixels. Inliers are defined as end-point error $<$ 3 pixels or 5\%. $^{\dag}$Using additional training sets. $^{\ddag}$A multi-frame method.)} \label{tab:results}
\scalebox{0.74}{
\begin{tabular}{|l|c c|c c|c c c|c c c c|}
\hline
\multirow{1}{*}{Method}   	                             	
&\multicolumn{2}{c|}{Sintel Clean} 							
&\multicolumn{2}{c|}{Sintel Final}						
&\multicolumn{3}{c|}{KITTI 2012} 
&\multicolumn{4}{c|}{KITTI 2015} \\

\multirow{1}{*}{}
&\multicolumn{1}{c}{train}&\multicolumn{1}{c|}{test}
&\multicolumn{1}{c}{train}&\multicolumn{1}{c|}{test}
&\multicolumn{1}{c}{train}&\multicolumn{1}{c}{test (All)}&\multicolumn{1}{c|}{test (Noc)}
&\multicolumn{1}{c}{train}&\multicolumn{1}{c}{train (Fl-all)}&\multicolumn{1}{c}{test (Fl-fg)}&\multicolumn{1}{c|}{test (Fl-all)} \\	
\hline

\multirow{1}{*}{FlowNetS~\cite{Dosovitskiy15}}				
&(3.66)&\multicolumn{1}{c|}{6.96}	           
&(4.44)&\multicolumn{1}{c|}{7.76}
&7.52&\multicolumn{1}{c}{9.1}&\multicolumn{1}{c|}{-}
&-&\multicolumn{1}{c}{-}&\multicolumn{1}{c}{-}&\multicolumn{1}{c|}{-} \\ 
                                
\multirow{1}{*}{FlowNetC~\cite{Dosovitskiy15}}				
&(3.78)&\multicolumn{1}{c|}{6.85}	           
&(5.28)&\multicolumn{1}{c|}{8.51}
&8.79&\multicolumn{1}{c}{-}&\multicolumn{1}{c|}{-}	
&-&\multicolumn{1}{c}{-}&\multicolumn{1}{c}{-}&\multicolumn{1}{c|}{-}\\
                                                                                            
\multirow{1}{*}{FlowNet2~\cite{Ilg17}}				
&(1.45)&\multicolumn{1}{c|}{4.16}	           
&(2.19)&\multicolumn{1}{c|}{5.74}
&(1.43)&\multicolumn{1}{c}{1.8}&\multicolumn{1}{c|}{1.0}
&(2.36)&\multicolumn{1}{c}{(8.88\%)}&\multicolumn{1}{c}{8.75\%}&\multicolumn{1}{c|}{11.48\%}\\ 

\multirow{1}{*}{FlowNet3~\cite{Ilg18}}				
&(1.47)&\multicolumn{1}{c|}{4.35}	           
&(2.12)&\multicolumn{1}{c|}{5.67}
&(1.19)&\multicolumn{1}{c}{-}&\multicolumn{1}{c|}{-}
&(1.79)&\multicolumn{1}{c}{-}&\multicolumn{1}{c}{-}&\multicolumn{1}{c|}{8.60\%}\\ 

\multirow{1}{*}{SPyNet~\cite{Ranjan17}}				
&(3.17)&\multicolumn{1}{c|}{6.64}	           
&(4.32)&\multicolumn{1}{c|}{8.36}
&3.36&\multicolumn{1}{c}{4.1}&\multicolumn{1}{c|}{2.0}	
&-&\multicolumn{1}{c}{-}&\multicolumn{1}{c}{43.62\%}&\multicolumn{1}{c|}{35.07\%}	\\  

\multirow{1}{*}{Devon~\cite{Lu20}}				
&-&\multicolumn{1}{c|}{4.34}           
&-&\multicolumn{1}{c|}{6.35}
&-&\multicolumn{1}{c}{2.6}&\multicolumn{1}{c|}{1.3}		
&-&\multicolumn{1}{c}{-}&\multicolumn{1}{c}{19.49\%}&\multicolumn{1}{c|}{14.31\%}\\

\multirow{1}{*}{PWC-Net~\cite{Sun18}}				
&(2.02)&\multicolumn{1}{c|}{4.39}	           
&(2.08)&\multicolumn{1}{c|}{5.04}
&(1.45)&\multicolumn{1}{c}{1.7}&\multicolumn{1}{c|}{0.9}
&(2.16)&\multicolumn{1}{c}{(9.80\%)}&\multicolumn{1}{c}{9.31\%}&\multicolumn{1}{c|}{9.60\%}\\  

\multirow{1}{*}{PWC-Net+~\cite{Sun19}}				
&(1.71)&\multicolumn{1}{c|}{3.45}          
&(2.34)&\multicolumn{1}{c|}{4.60}
&(0.99)&\multicolumn{1}{c}{\underline{1.4}}&\multicolumn{1}{c|}{\underline{0.8}}	
&(1.47)&\multicolumn{1}{c}{(7.59\%)}&\multicolumn{1}{c}{7.88\%}&\multicolumn{1}{c|}{7.72\%}\\ 
 
\multirow{1}{*}{IRR-PWC~\cite{Hur19}}				
&(1.92)&\multicolumn{1}{c|}{3.84}           
&(2.51)&\multicolumn{1}{c|}{4.58}
&-&\multicolumn{1}{c}{1.6}&\multicolumn{1}{c|}{0.9}		
&(1.63)&\multicolumn{1}{c}{(5.32\%)}&\multicolumn{1}{c}{\underline{7.52\%}}&\multicolumn{1}{c|}{7.65\%}\\
                               
\multirow{1}{*}{SENSE~\cite{Jiang19}$^{\dag}$}				
&(1.54)&\multicolumn{1}{c|}{3.60}           
&(2.05)&\multicolumn{1}{c|}{4.86}
&(1.18)&\multicolumn{1}{c}{1.5}&\multicolumn{1}{c|}{-}		
&(2.05)&\multicolumn{1}{c}{(9.69\%)}&\multicolumn{1}{c}{9.33\%}&\multicolumn{1}{c|}{8.16\%}\\
                                                          
\multirow{1}{*}{HD$^{3}$~\cite{Yin19}$^{\dag}$}				
&(1.70)&\multicolumn{1}{c|}{4.79}           
&(1.17)&\multicolumn{1}{c|}{4.67}
&(0.81)&\multicolumn{1}{c}{1.4}&\multicolumn{1}{c|}{\textbf{0.7}}		
&(1.31)&\multicolumn{1}{c}{(4.10\%)}&\multicolumn{1}{c}{9.02\%}&\multicolumn{1}{c|}{\textbf{6.55\%}}\\

\multirow{1}{*}{SelFlow~\cite{Liu19}$^{\dag,\ddag}$}				
&(1.68)&\multicolumn{1}{c|}{3.75}           
&(1.77)&\multicolumn{1}{c|}{\textbf{4.26}}
&(0.76)&\multicolumn{1}{c}{1.5}&\multicolumn{1}{c|}{0.9}		
&(1.18)&\multicolumn{1}{c}{-}&\multicolumn{1}{c}{12.48\%}&\multicolumn{1}{c|}{{8.42\%}}\\

\multirow{1}{*}{VCN-small~\cite{Yang19}}				
&(1.84)&\multicolumn{1}{c|}{3.26}           
&(2.44)&\multicolumn{1}{c|}{4.73}
&-&\multicolumn{1}{c}{-}&\multicolumn{1}{c|}{-}		
&(1.41)&\multicolumn{1}{c}{(5.5\%)}&\multicolumn{1}{c}{-}&\multicolumn{1}{c|}{{7.74\%}}\\

\multirow{1}{*}{LiteFlowNet~\cite{Hui18}}				
&(1.35)&\multicolumn{1}{c|}{4.54}           
&(1.78)&\multicolumn{1}{c|}{5.38}
&(1.05)&\multicolumn{1}{c}{1.6}&\multicolumn{1}{c|}{\underline{0.8}}	
&(1.62)&\multicolumn{1}{c}{(5.58\%)}&\multicolumn{1}{c}{7.99\%}&\multicolumn{1}{c|}{9.38\%}\\    

\multirow{1}{*}{LiteFlowNet2~\cite{Hui20}}				
&(1.30)&\multicolumn{1}{c|}{3.48} 
&(1.62)&\multicolumn{1}{c|}{4.69}
&(0.95)&\multicolumn{1}{c}{\underline{1.4}}&\multicolumn{1}{c|}{\textbf{0.7}}		
&(1.33)&\multicolumn{1}{c}{(4.32\%)}&\multicolumn{1}{c}{7.64\%}&\multicolumn{1}{c|}{7.62\%}\\ 

\multirow{1}{*}{\textbf{LiteFlowNet3}}				
&(1.32)&\multicolumn{1}{c|}{\textbf{2.99}}           
&(1.76)&\multicolumn{1}{c|}{\underline{4.45}}
&(0.91)&\multicolumn{1}{c}{\textbf{1.3}}&\multicolumn{1}{c|}{\textbf{0.7}}		
&\multicolumn{1}{c}{(1.26)}&\multicolumn{1}{c}{(3.82\%)}&\multicolumn{1}{c}{7.75\%}&\multicolumn{1}{c|}{7.34\%}\\ 

\multirow{1}{*}{\textbf{LiteFlowNet3-S}}				
&(1.43)&\multicolumn{1}{c|}{\underline{3.03}}           
&(1.90)&\multicolumn{1}{c|}{4.53}
&\multicolumn{1}{c}{(0.94)}&\multicolumn{1}{c}{\textbf{1.3}}&\multicolumn{1}{c|}{\textbf{0.7}}	
&\multicolumn{1}{c}{(1.39)}&\multicolumn{1}{c}{(4.35\%)}&\multicolumn{1}{c}{\textbf{6.96\%}}&\multicolumn{1}{c|}{\underline{7.22\%}}\\ 
\hline
\end{tabular}}
\end{table*}

\begin{table}[t]
\centering
\caption{AEE results on the testing sets of Sintel. (Note: $^{\dag}$Using additional training sets.)} \label{tab:detailed sintel}
\scalebox{1}{
\begin{tabular}{|l|cc|cc|cc|}
\hline
\multicolumn{1}{|l|}{Models}
&\multicolumn{2}{c|}{All}
&\multicolumn{2}{c|}{Matched}
&\multicolumn{2}{c|}{Unmatched} \\

\multicolumn{1}{|l|}{}
&Clean&\multicolumn{1}{c|}{Final}
&Clean&\multicolumn{1}{c|}{Final}
&Clean&\multicolumn{1}{c|}{Final} \\
\hline
\multicolumn{1}{|l|}{FlowNet2~\cite{Ilg17}}
&4.16&\multicolumn{1}{c|}{5.74}
&1.56&\multicolumn{1}{c|}{2.75}
&25.40&\multicolumn{1}{c|}{30.11} \\

\multicolumn{1}{|l|}{Devon~\cite{Lu20}}
&4.34&\multicolumn{1}{c|}{6.35}
&1.74&\multicolumn{1}{c|}{3.23}
&25.58&\multicolumn{1}{c|}{31.78} \\

\multicolumn{1}{|l|}{PWC-Net+~\cite{Sun19}}
&3.45&\multicolumn{1}{c|}{4.60}
&1.41&\multicolumn{1}{c|}{2.25}
&20.12&\multicolumn{1}{c|}{23.70} \\

\multicolumn{1}{|l|}{IRR-PWC~\cite{Hur19}}
&3.84&\multicolumn{1}{c|}{4.58}
&1.47&\multicolumn{1}{c|}{2.15}
&23.22&\multicolumn{1}{c|}{24.36} \\

\multicolumn{1}{|l|}{SENSE~\cite{Jiang19}$^{\dag}$}
&3.60&\multicolumn{1}{c|}{4.86}
&1.38&\multicolumn{1}{c|}{2.30}
&21.75&\multicolumn{1}{c|}{25.73} \\

\multicolumn{1}{|l|}{HD$^{3}$~\cite{Yin19}$^{\dag}$}
&4.79&\multicolumn{1}{c|}{4.67}
&1.62&\multicolumn{1}{c|}{2.17}
&30.63&\multicolumn{1}{c|}{24.99} \\

\multicolumn{1}{|l|}{LiteFlowNet2~\cite{Hui20}}
&3.48&\multicolumn{1}{c|}{4.69}
&1.33&\multicolumn{1}{c|}{2.25}
&20.64&\multicolumn{1}{c|}{24.57} \\

\multicolumn{1}{|l|}{\textbf{LiteFlowNet3}}
&\textbf{2.99}&\multicolumn{1}{c|}{\textbf{4.45}}
&\textbf{1.15}&\multicolumn{1}{c|}{\textbf{2.09}}
&\textbf{18.08}&\multicolumn{1}{c|}{\textbf{23.68}} \\
\hline 
\end{tabular}}
\end{table}

\begin{figure*}[t]
\centering
\captionsetup[subfigure]{labelformat=empty, justification=centering}
\captionsetup[subfloat]{farskip=0pt,captionskip=0pt}
\begin{tabular}{ccccc}
   \includegraphics[width=2.45cm]{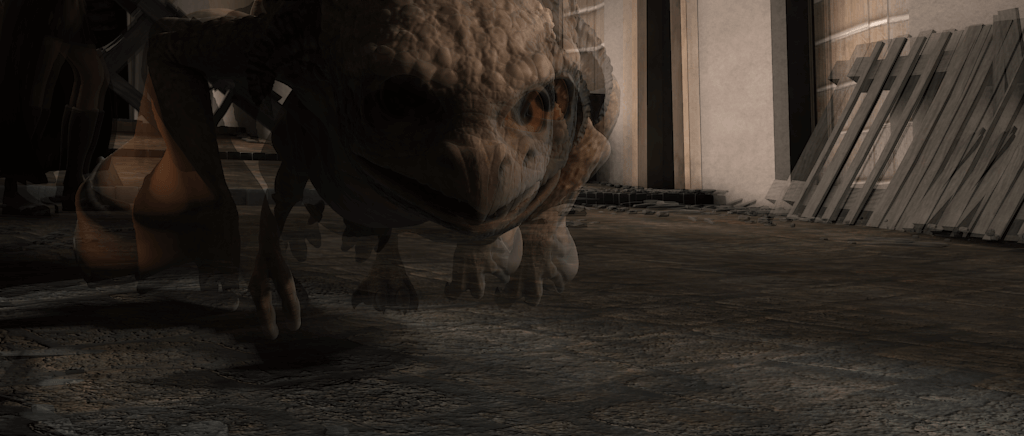}\hfill
   \includegraphics[width=2.45cm]{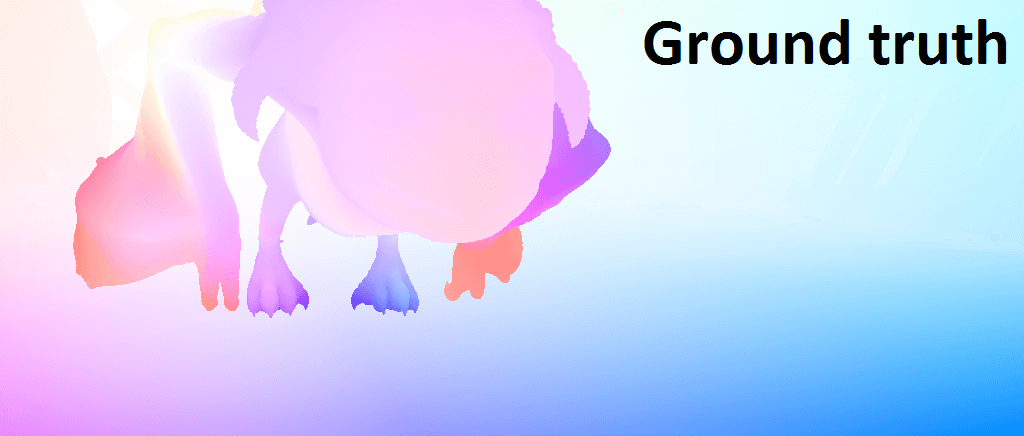}\hfill
   \includegraphics[width=2.45cm]{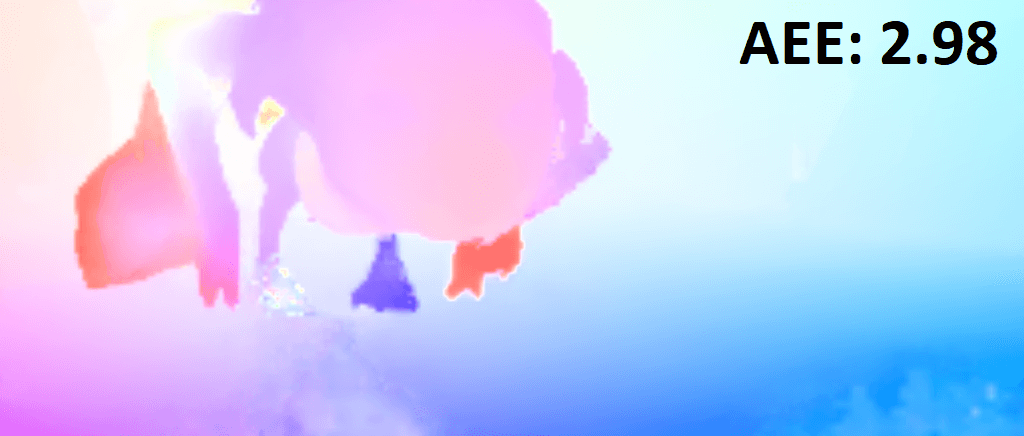}\hfill
   \includegraphics[width=2.45cm]{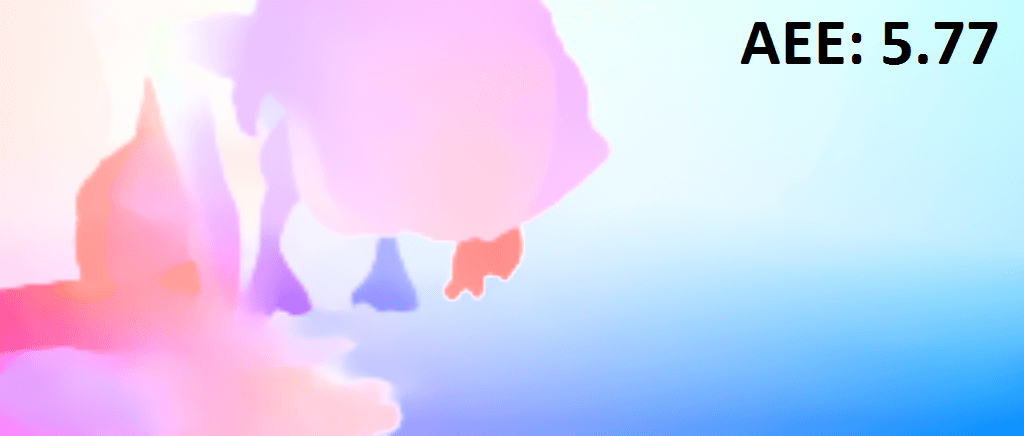}\hfill
   \includegraphics[width=2.45cm]{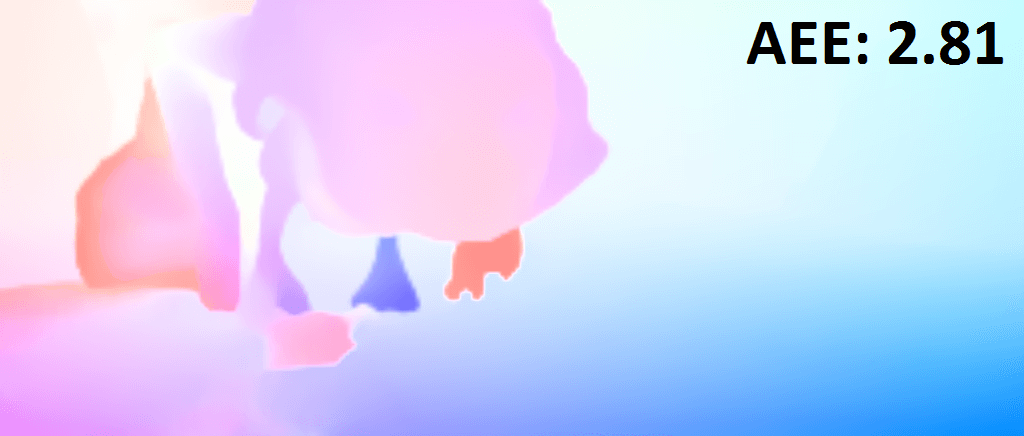}\\

   \includegraphics[width=2.45cm]{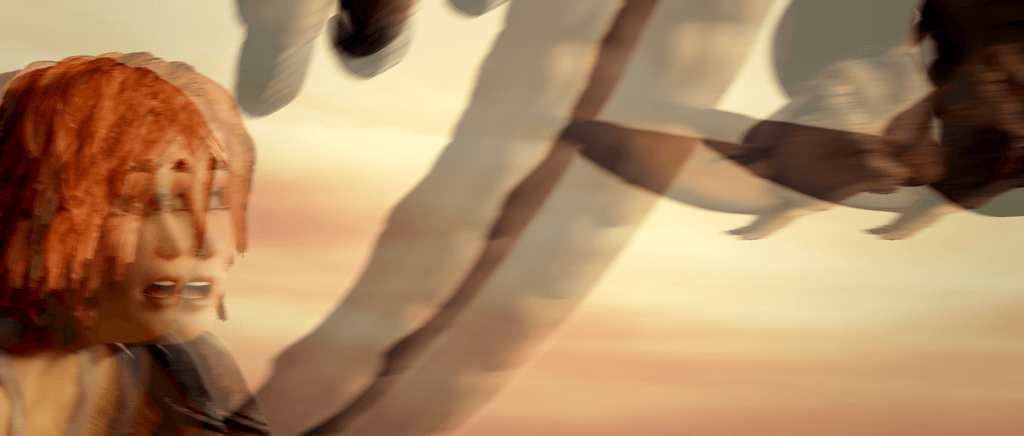}\hfill
   \includegraphics[width=2.45cm]{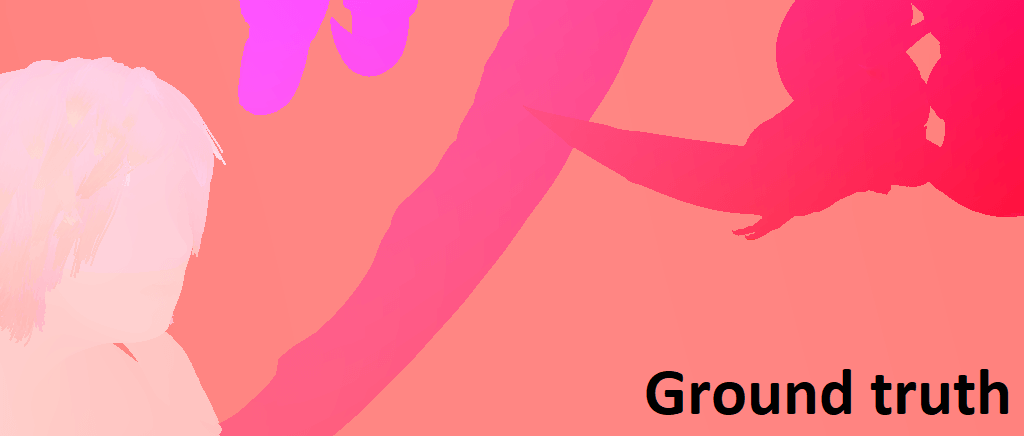}\hfill
   \includegraphics[width=2.45cm]{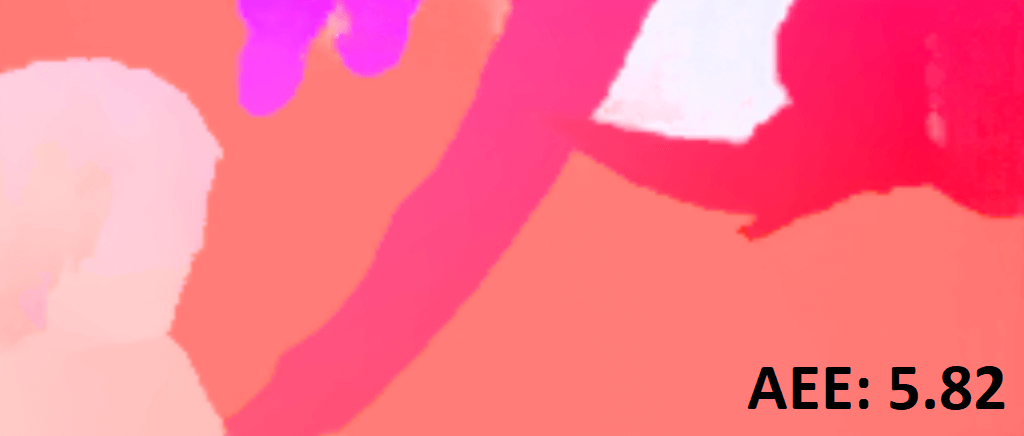}\hfill
   \includegraphics[width=2.45cm]{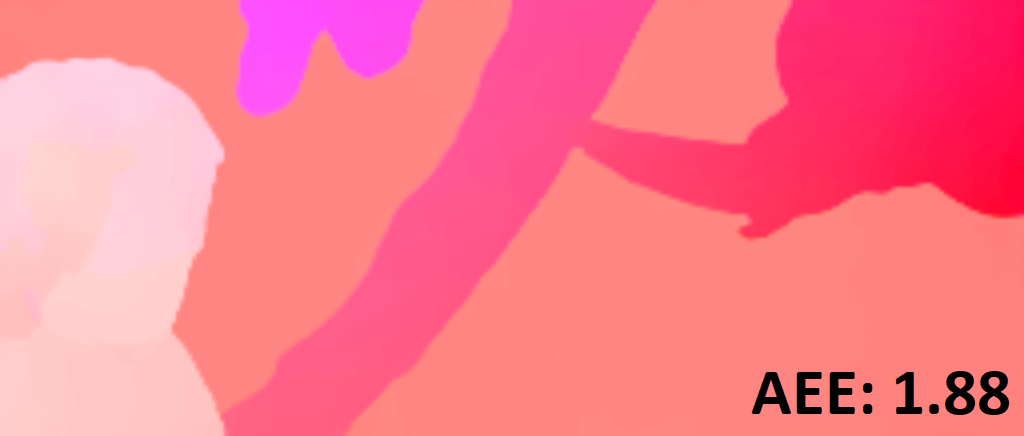}\hfill
   \includegraphics[width=2.45cm]{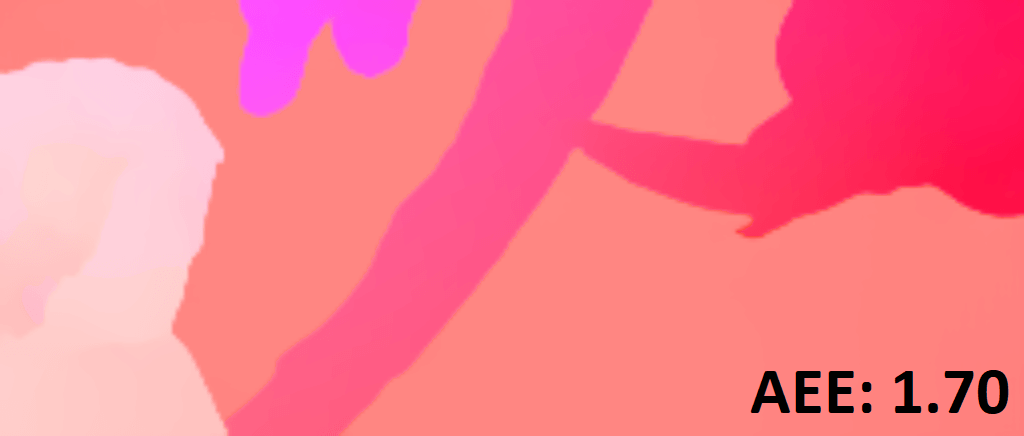} \\

   \includegraphics[width=2.45cm]{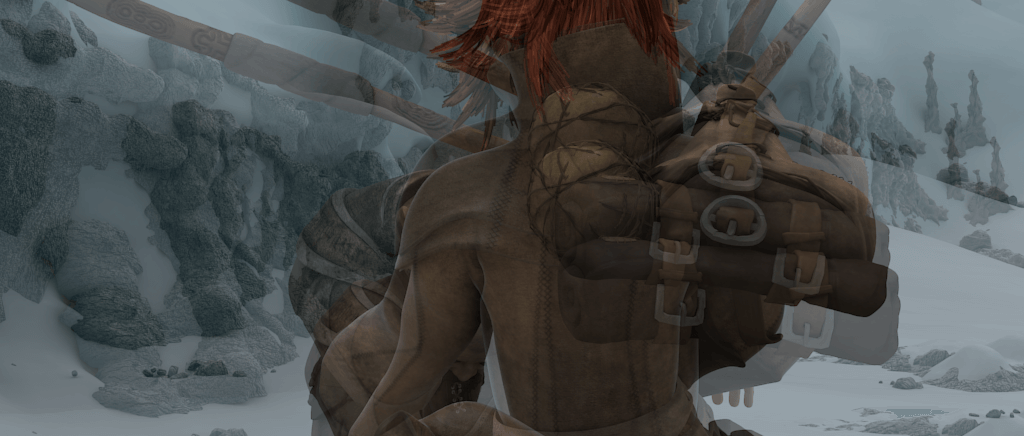}\hfill
   \includegraphics[width=2.45cm]{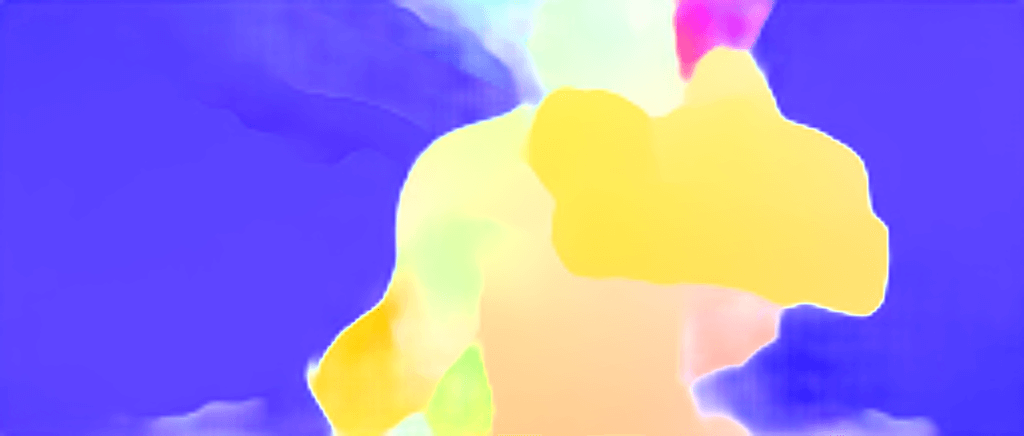}\hfill
   \includegraphics[width=2.45cm]{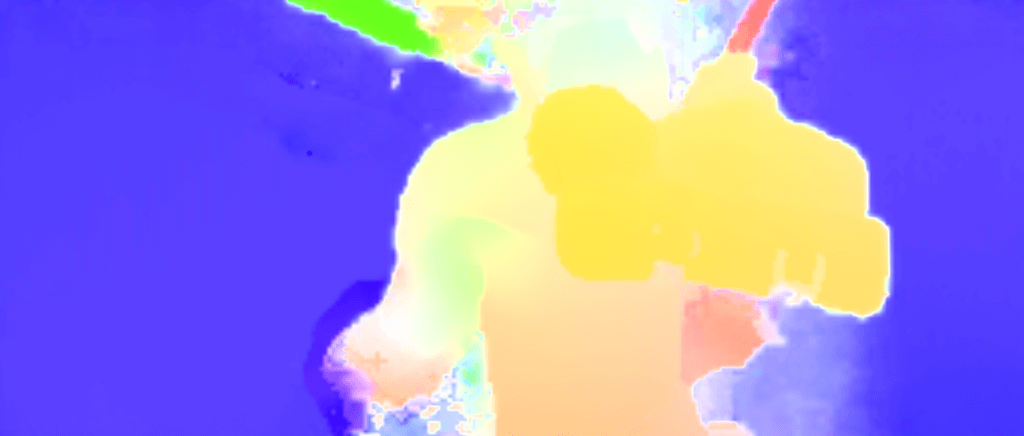}\hfill
   \includegraphics[width=2.45cm]{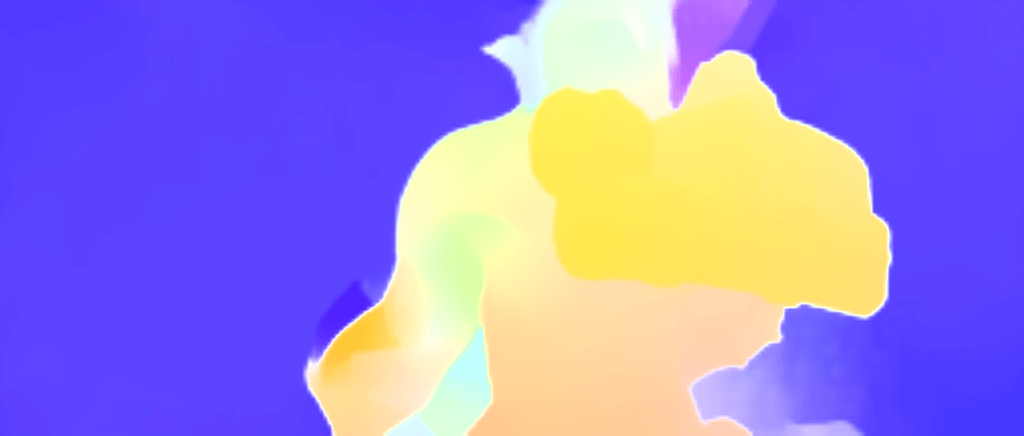}\hfill
   \includegraphics[width=2.45cm]{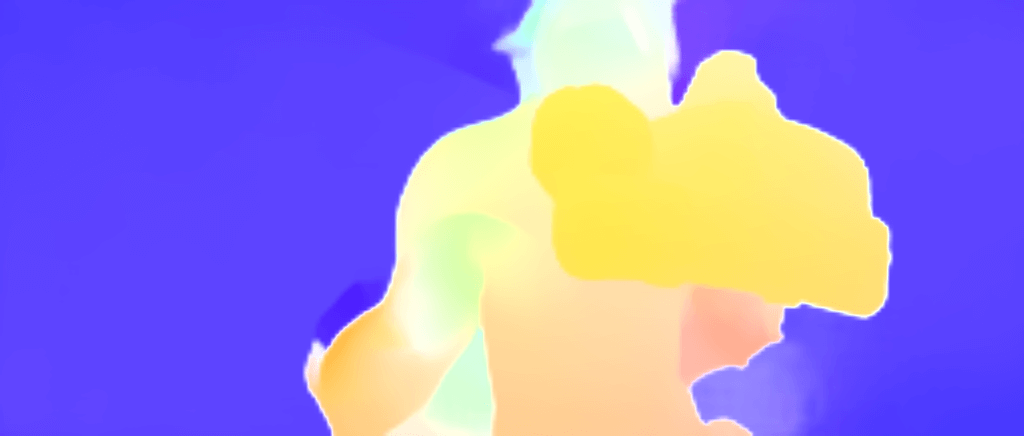} \\

   \subfloat[Image overlay]{\includegraphics[width=2.45cm]{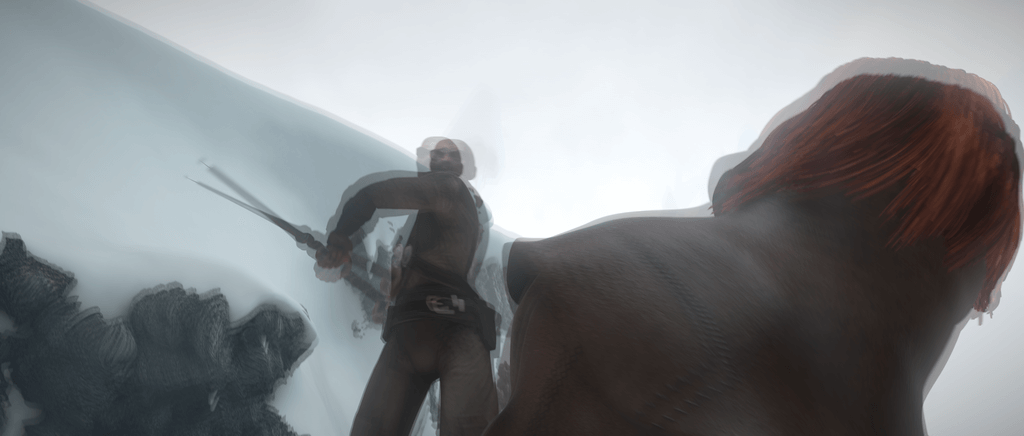}}\hfill
   \subfloat[PWC-Net+~\cite{Sun19}]{\includegraphics[width=2.45cm]{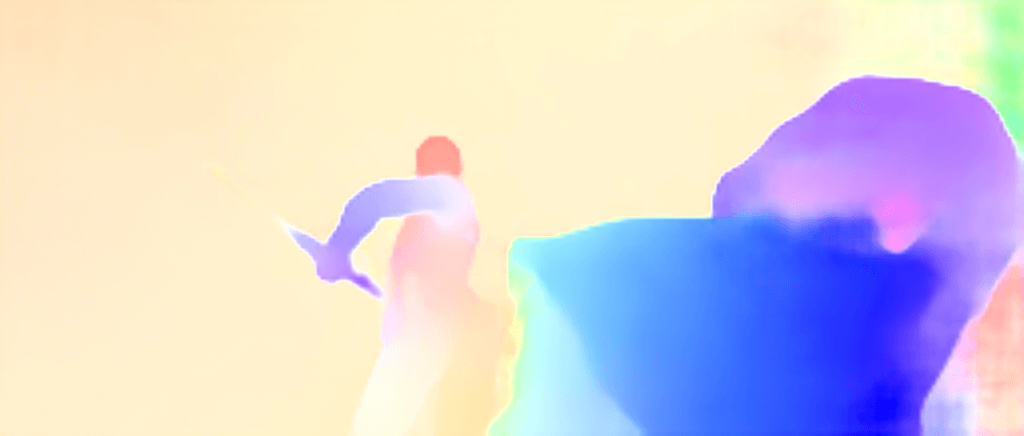}}\hfill
   \subfloat[HD$^{3}$~\cite{Yin19}]{\includegraphics[width=2.45cm]{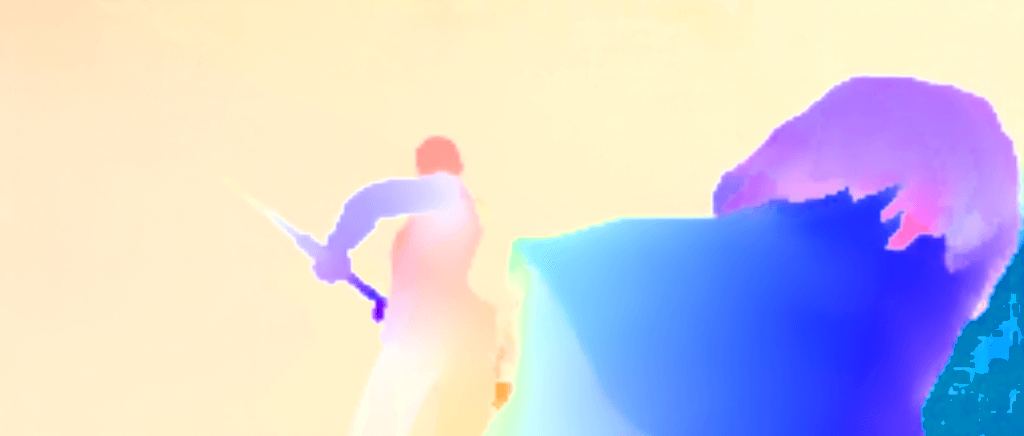}}\hfill
   \subfloat[LiteFlowNet2\cite{Hui20}]{\includegraphics[width=2.45cm]{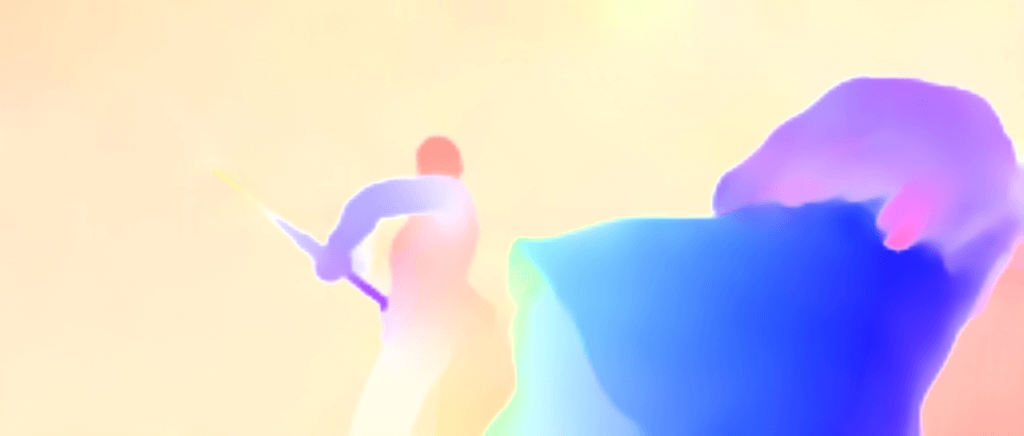}}\hfill
   \subfloat[LiteFlowNet3]{\includegraphics[width=2.45cm]{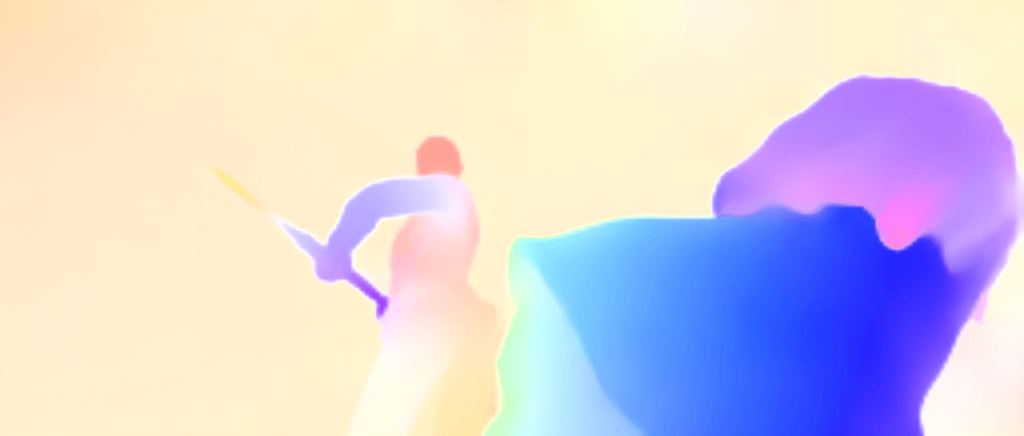}} \\
\end{tabular}
\caption{Examples of flow fields on Sintel training set (Clean pass: first row, Final pass: second row) and testing set (Clean pass: third row, Final pass: forth row).}
\label{fig:sintel results}
\end{figure*}

\begin{figure*}[!]
\centering
\captionsetup[subfigure]{labelformat=empty, justification=centering}
\captionsetup[subfloat]{farskip=0pt,captionskip=0pt}
\begin{tabular}{cccc}
   \includegraphics[width=3.05cm]{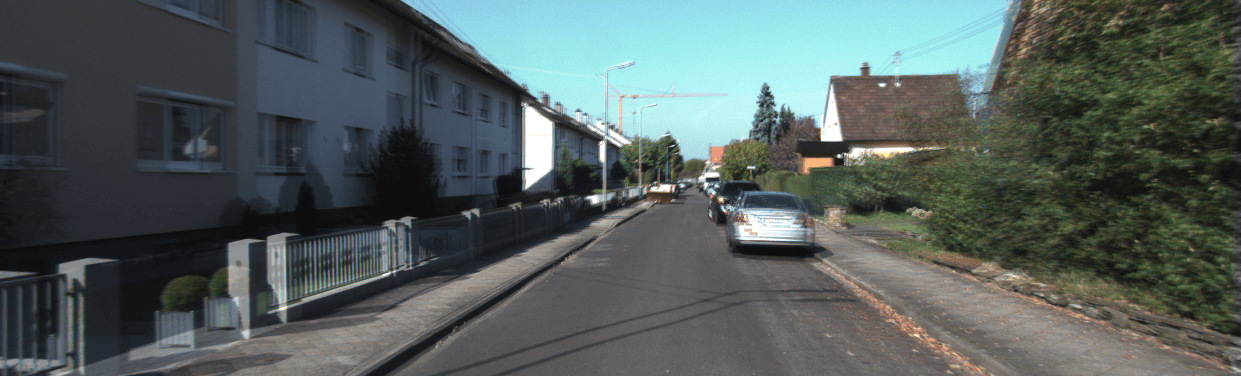}\hfill
   \includegraphics[width=3.05cm]{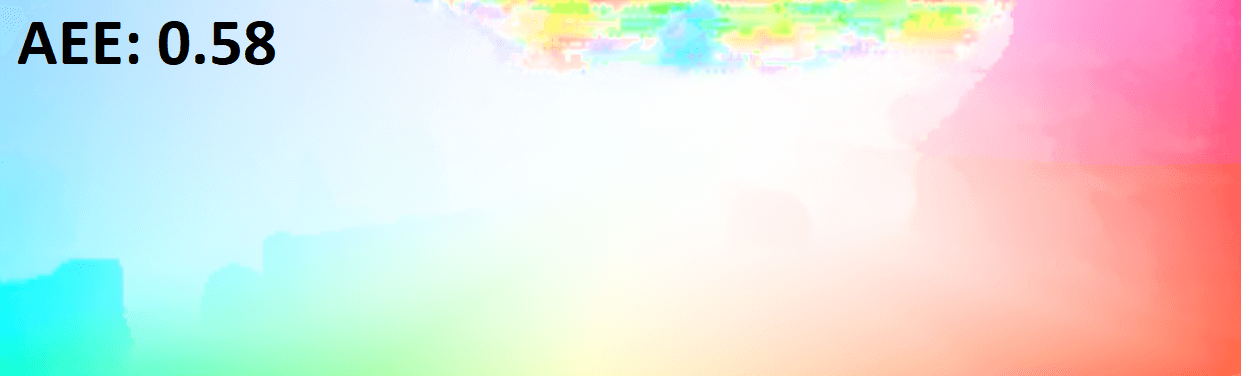}\hfill
   \includegraphics[width=3.05cm]{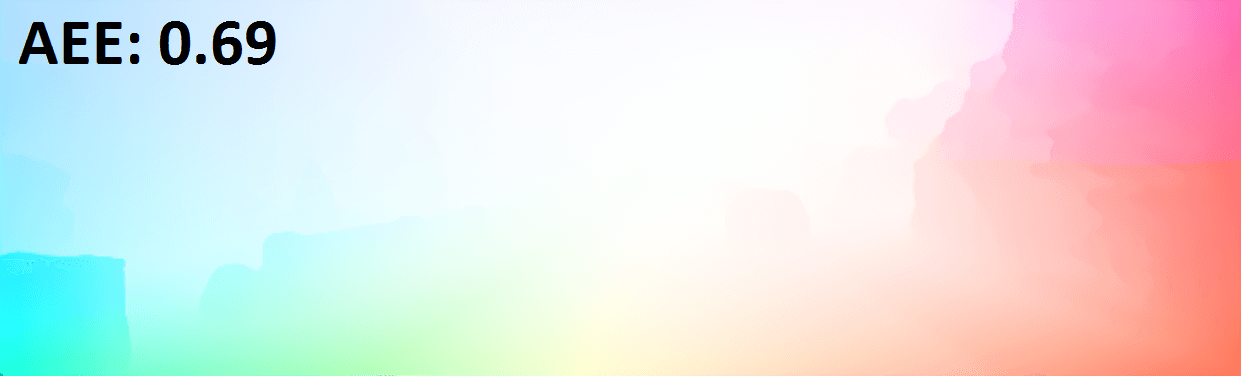}\hfill
   \includegraphics[width=3.05cm]{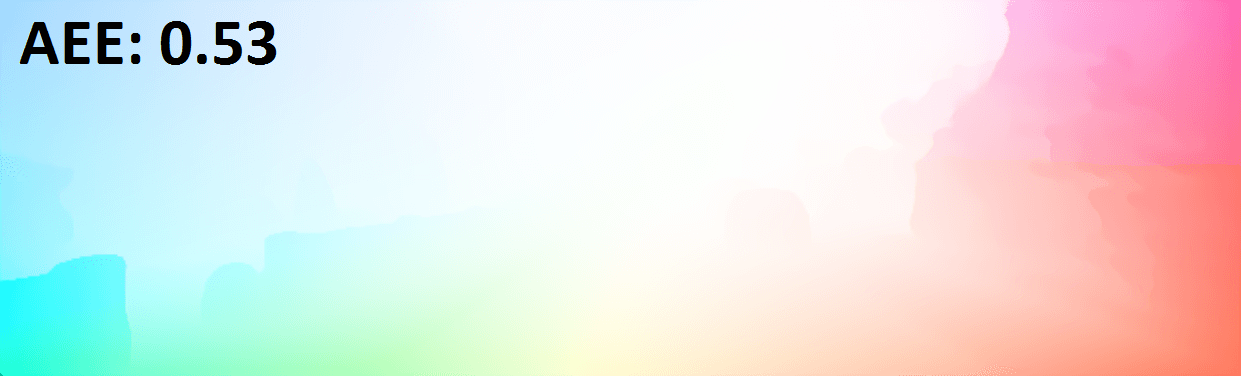} \\

   \includegraphics[width=3.05cm]{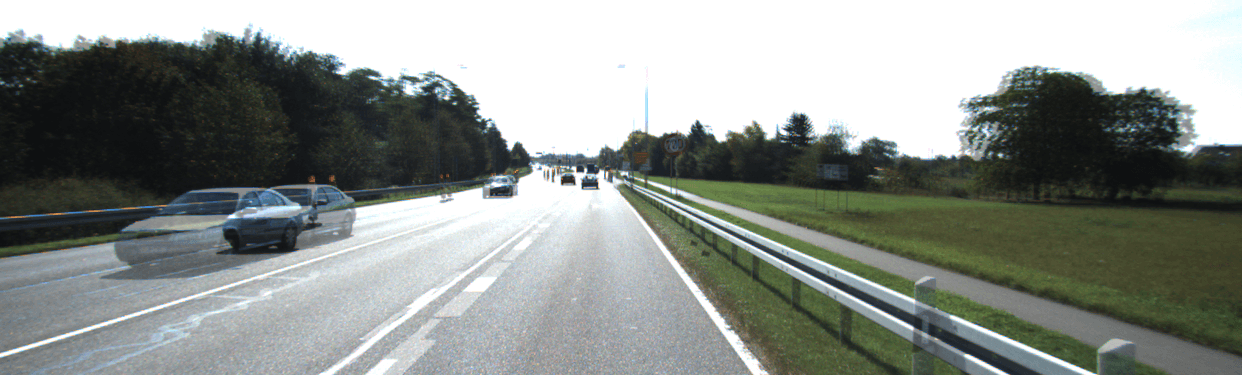}\hfill
   \includegraphics[width=3.05cm]{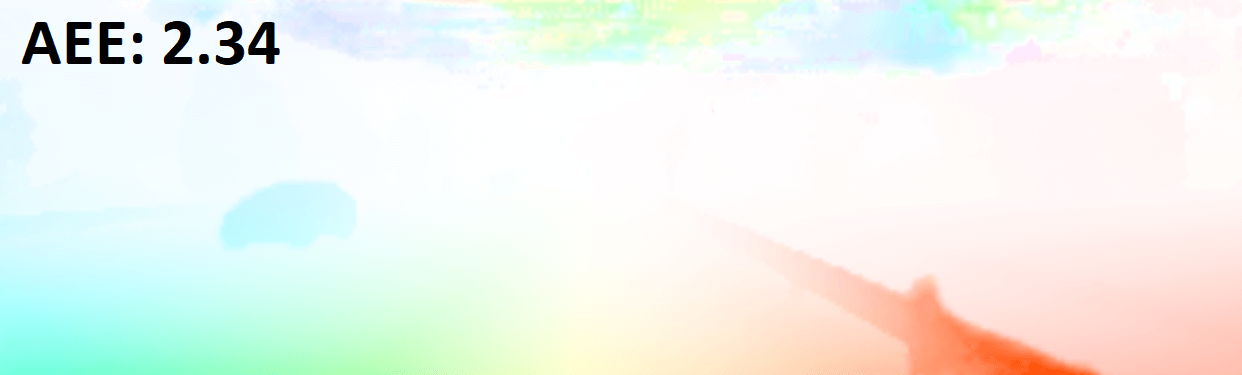}\hfill
   \includegraphics[width=3.05cm]{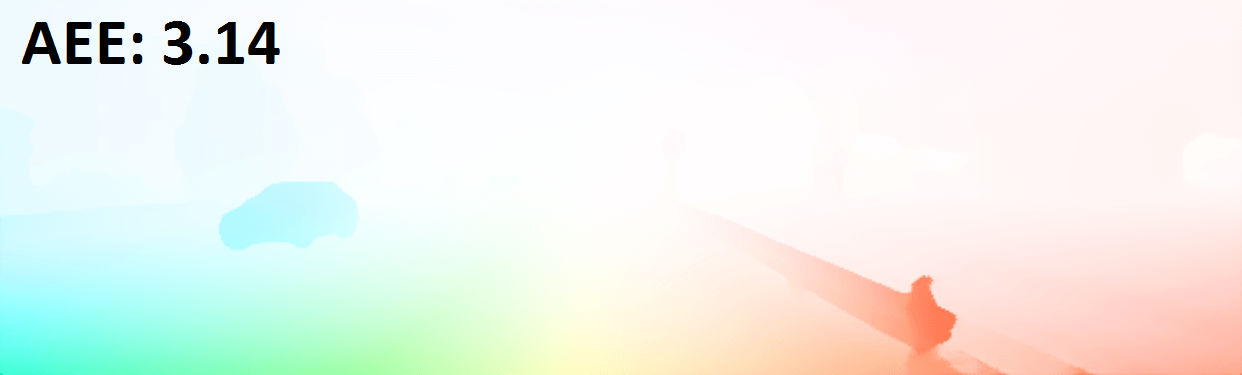}\hfill
   \includegraphics[width=3.05cm]{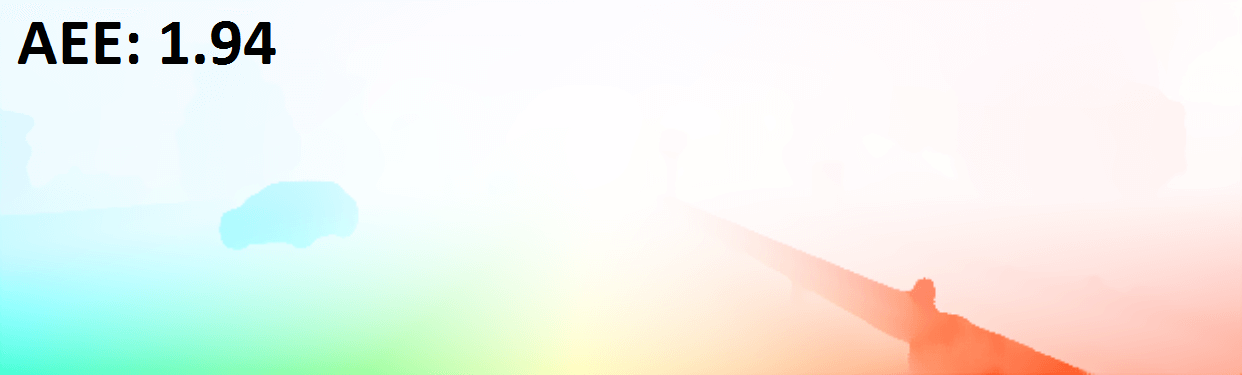} \\

   \includegraphics[width=3.05cm]{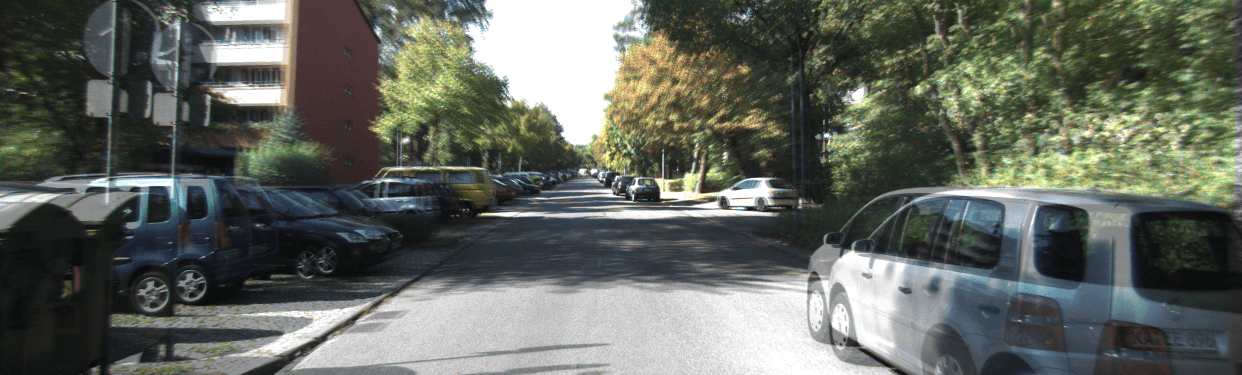}\hfill
   \includegraphics[width=3.05cm]{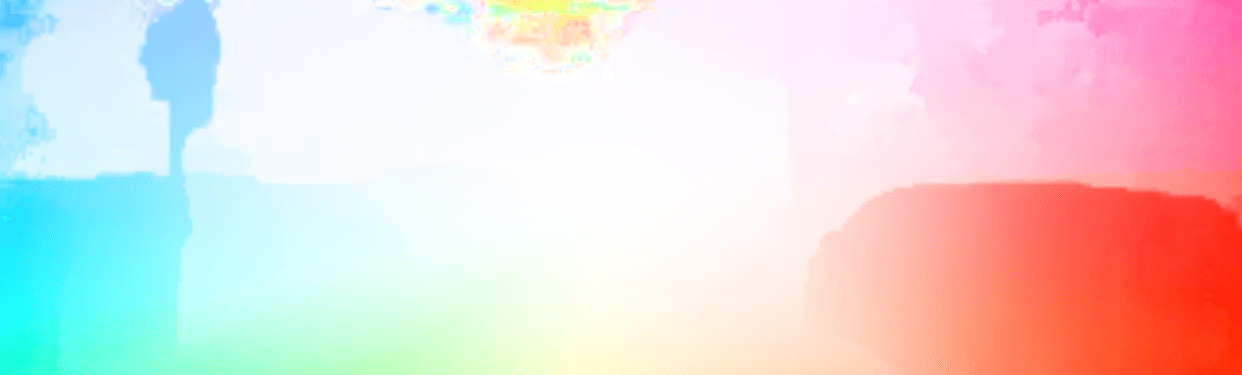}\hfill
   \includegraphics[width=3.05cm]{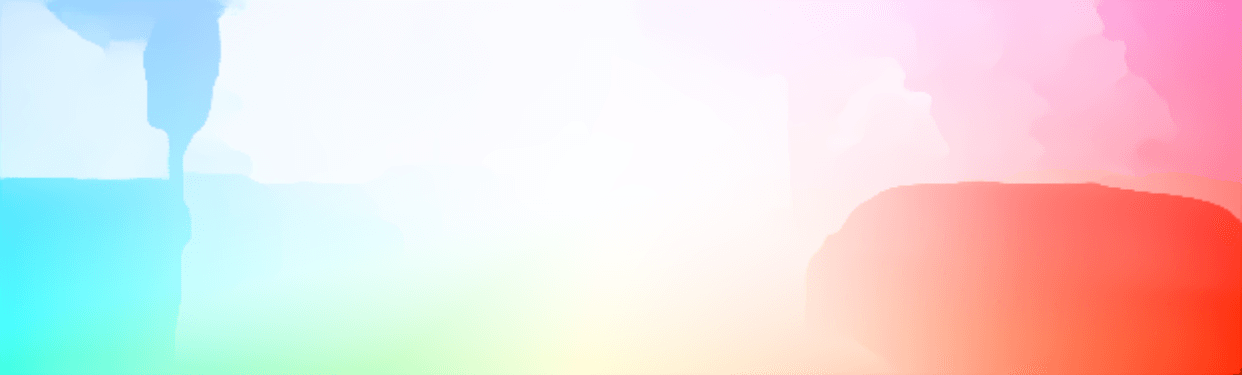}\hfill
   \includegraphics[width=3.05cm]{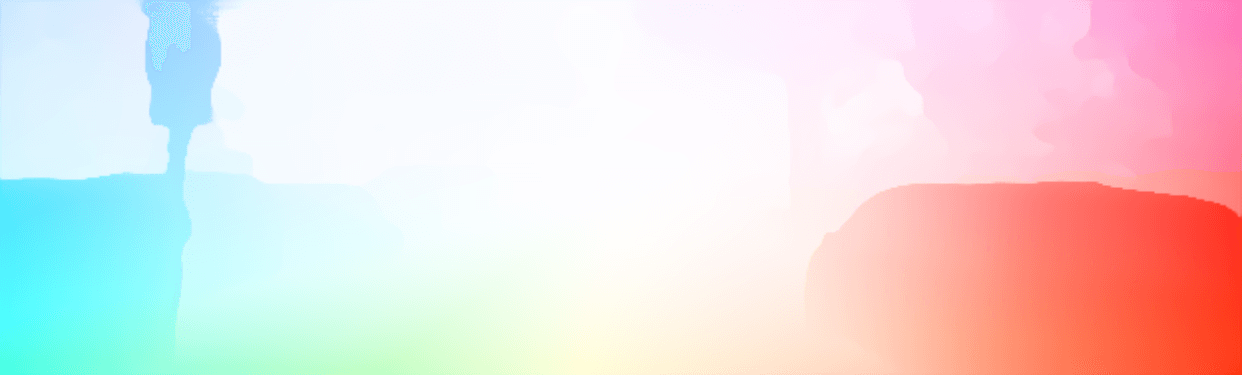} \\

   \subfloat[Image overlay]{\includegraphics[width=3.05cm]{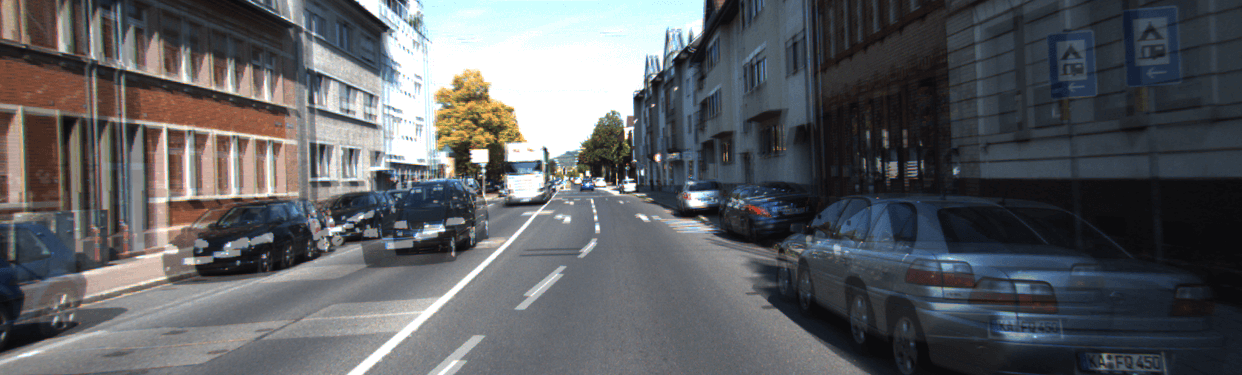}}\hfill
   \subfloat[HD$^{3}$~\cite{Yin19}]{\includegraphics[width=3.05cm]{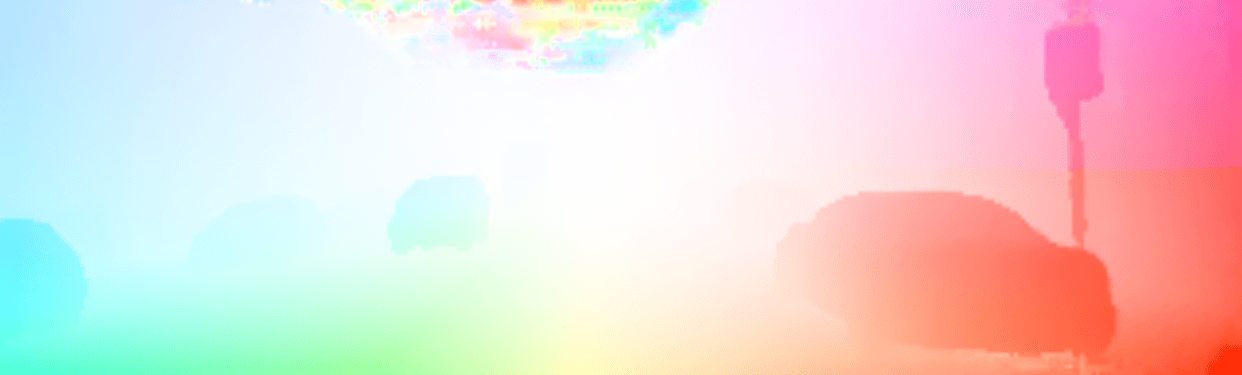}}\hfill
   \subfloat[LiteFlowNet2~\cite{Hui20}]{\includegraphics[width=3.05cm]{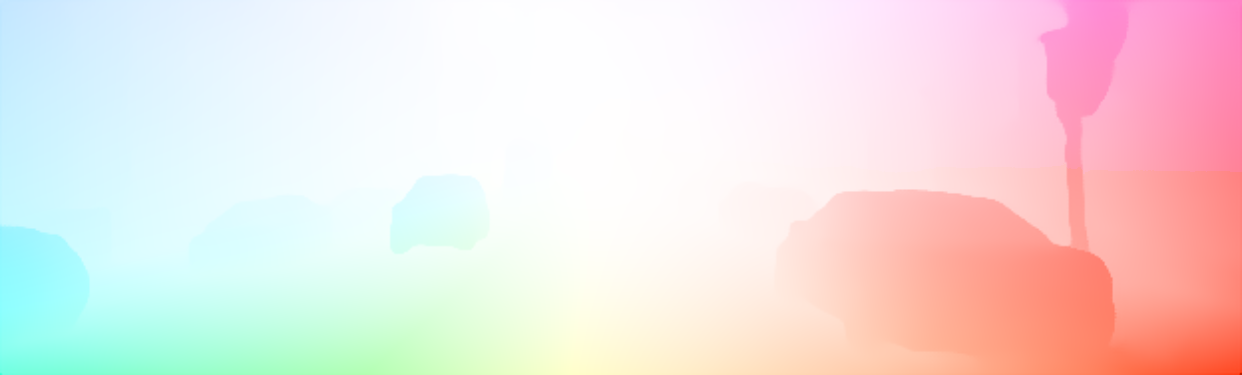}}\hfill
   \subfloat[LiteFlowNet3]{\includegraphics[width=3.05cm]{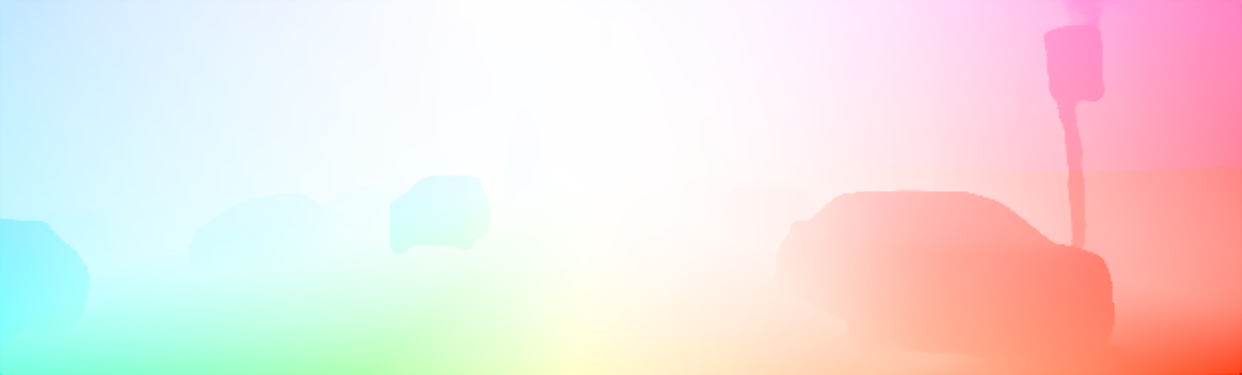}} \\
\end{tabular}
\caption{Examples of flow fields on KITTI training set (2012: first row, 2015: second row) and testing set (2012: third row, 2015: fourth row).}
\label{fig:kitti results}
\end{figure*}

\noindent\textbf{Qualitative Results.} Examples of optical flow predictions on Sintel and KITTI are shown in Figs.~\ref{fig:sintel results} and \ref{fig:kitti results}, respectively. AEE evaluated on the respective training sets is also provided. 
For Sintel, the flow fields resulting from LiteFlowNet3 contain less artifacts when comparing with the other state-of-the-art methods.
As shown in the second row of Fig.~\ref{fig:kitti results}, a portion of optical flow over the road fence cannot be recovered by LiteFlowNet2~\cite{Hui20}. On the contrary, it is fully recovered by HD$^{3}$~\cite{Yin19} and LiteFlowNet3.
Flow bleeding is observed over the road signs for LiteFlowNet2 as illustrated in the third and fourth rows of Fig.~\ref{fig:kitti results} while HD$^{3}$ and LiteFlowNet3 do not have such a problem.
Despite HD$^{3}$ is pre-trained on ImageNet and uses 7.7 times more model parameters than LiteFlowNet3, there are serious artifacts on the generated flow fields as shown in the second column of Fig.~\ref{fig:kitti results}. 
The above observations suggest that LiteFlowNet3 incorporating the cost volume modulation and flow field deformation is effective in generating optical flow with high accuracy and less artifacts.

\noindent\textbf{Runtime and Model Size.} We measure runtime using a Sintel image pair ($1024\times436$) on a machine equipped with Intel Xeon E5 2.2GHz and NVIDIA GTX 1080. Timing is averaged over 100 runs. LiteFlowNet3 needs 59ms for computation and has 5.2M parameters. When weight sharing is not used, the model size is 7.5M. The runtimes of the state-of-the-art 2-frame methods HD$^{3}$~\cite{Yin19} and IRR-PWC~\cite{Hur19} are 128ms and 180ms, respectively. While HD$^{3}$ and IRR-PWC have 39.9M and 6.4M parameters, respectively. 

\subsection{Ablation Study} \label{sec:ablation study}
\begin{table}[t]
\centering
\caption{AEE results of variants of LiteFlowNet3 having some of the components disabled. (Note: The symbol ``-" indicates that confidence map is not being used.)} \label{tab:ablation study}
\scalebox{1}{
\begin{tabular}{|l|c|c|c|c|c|}
\hline
\multicolumn{1}{|l|}{Settings}
&\multicolumn{1}{>{\centering}m{1.2cm}|}{NO}
&\multicolumn{1}{>{\centering}m{1.2cm}|}{CM-}
&\multicolumn{1}{>{\centering}m{1.2cm}|}{CMFD-}
&\multicolumn{1}{>{\centering}m{1.2cm}|}{CM}
&\multicolumn{1}{c|}{~CMFD}  \\
\hline

\multicolumn{1}{|l|}{\textbf{C}ost Volume \textbf{M}odulation}
&\multicolumn{1}{c|}{\xmark}
&\multicolumn{1}{c|}{\cmark}
&\multicolumn{1}{c|}{\cmark}
&\multicolumn{1}{c|}{\cmark}
&\multicolumn{1}{c|}{\cmark} \\

\multicolumn{1}{|l|}{\textbf{F}low Field \textbf{D}eformation}
&\multicolumn{1}{c|}{\xmark}
&\multicolumn{1}{c|}{\xmark}
&\multicolumn{1}{c|}{\cmark}
&\multicolumn{1}{c|}{\xmark}
&\multicolumn{1}{c|}{\cmark} \\

\multicolumn{1}{|l|}{Confidence map}
&\multicolumn{1}{c|}{\xmark}
&\multicolumn{1}{c|}{\xmark}
&\multicolumn{1}{c|}{\xmark}
&\multicolumn{1}{c|}{\cmark}
&\multicolumn{1}{c|}{\cmark} \\
\hline\hline

\multicolumn{1}{|l|}{Sintel clean (training set)}
&\multicolumn{1}{c|}{2.78}
&\multicolumn{1}{c|}{2.66}
&\multicolumn{1}{c|}{2.63}
&\multicolumn{1}{c|}{2.65}
&\multicolumn{1}{c|}{\textbf{2.59}} \\

\multicolumn{1}{|l|}{Sintel final (training set)}
&\multicolumn{1}{c|}{4.14}
&\multicolumn{1}{c|}{4.09}
&\multicolumn{1}{c|}{4.06}
&\multicolumn{1}{c|}{4.02}
&\multicolumn{1}{c|}{\textbf{3.91}} \\

\multicolumn{1}{|l|}{KITTI 2012 (training set)}
&\multicolumn{1}{c|}{4.11}
&\multicolumn{1}{c|}{4.02}
&\multicolumn{1}{c|}{4.06}
&\multicolumn{1}{c|}{3.95}
&\multicolumn{1}{c|}{\textbf{3.88}} \\

\multicolumn{1}{|l|}{KITTI 2015 (training set)}
&\multicolumn{1}{c|}{11.31}
&\multicolumn{1}{c|}{11.01}
&\multicolumn{1}{c|}{10.97}
&\multicolumn{1}{c|}{10.65}
&\multicolumn{1}{c|}{\textbf{10.40}} \\
\hline 
\end{tabular}}
\end{table}

\begin{figure*}[t]
\centering
\captionsetup[subfigure]{labelformat=empty, justification=centering}
\captionsetup[subfloat]{farskip=0pt,captionskip=0pt}
\begin{tabular}{cccc}
   \includegraphics[width=3.05cm]{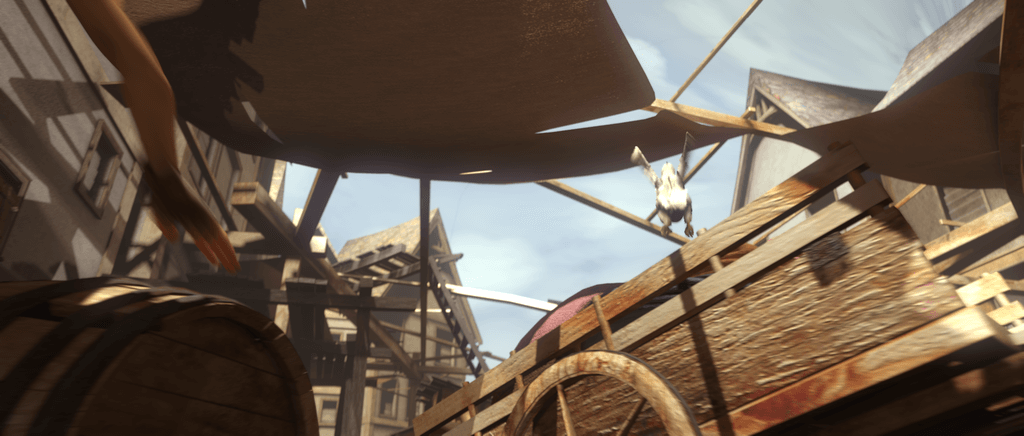}\hfill
   \includegraphics[width=3.05cm]{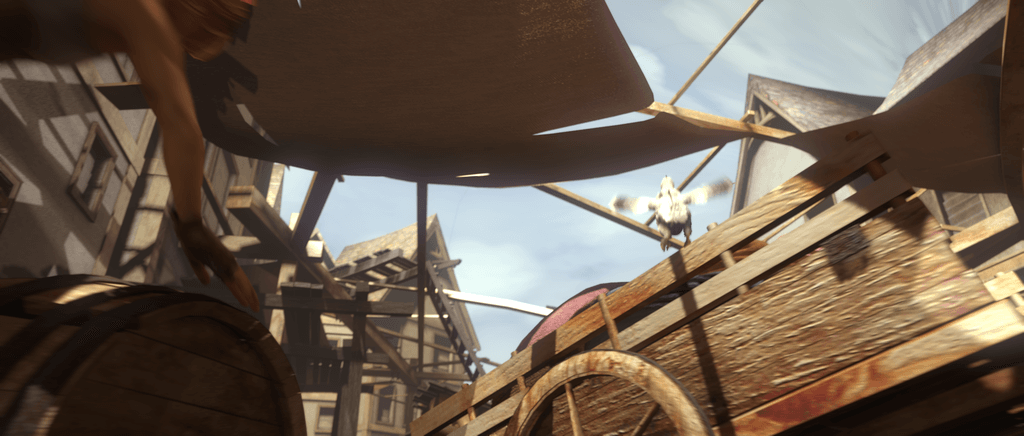}\hfill
   \includegraphics[width=3.05cm]{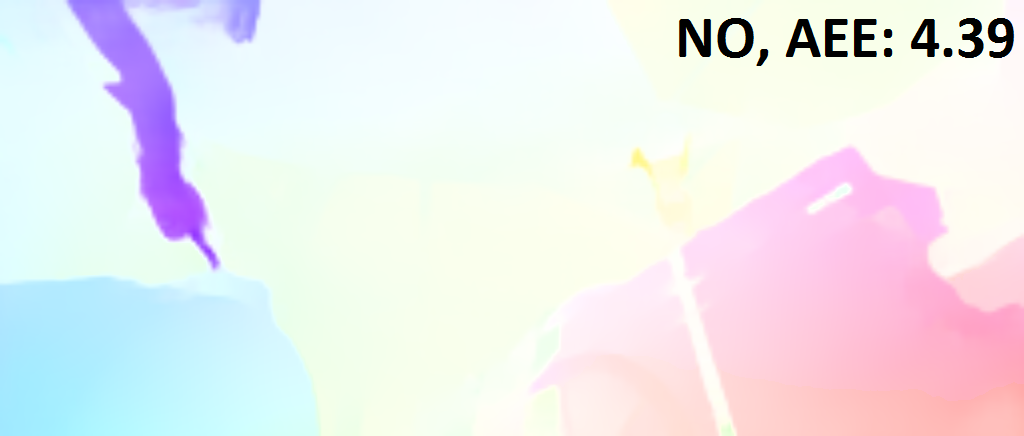}\hfill
   \includegraphics[width=3.05cm]{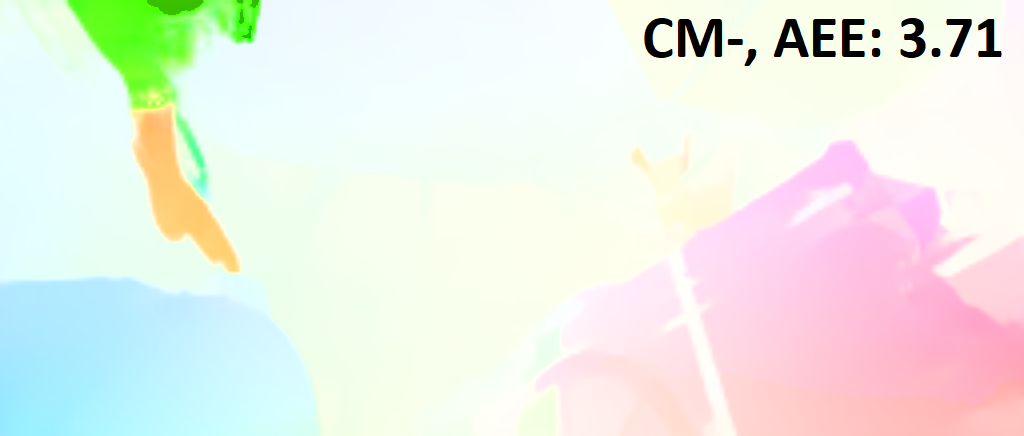}\vspace{-0.1cm}\\
	
   \includegraphics[width=3.05cm]{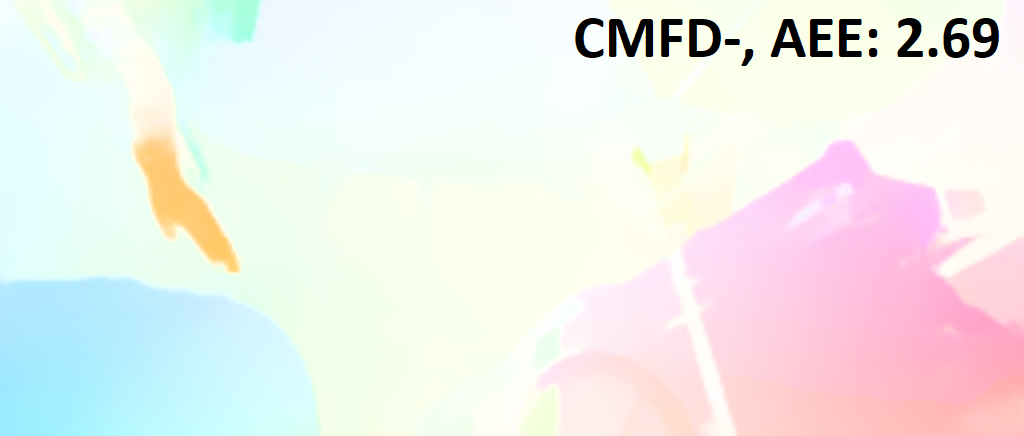}\hfill
   \includegraphics[width=3.05cm]{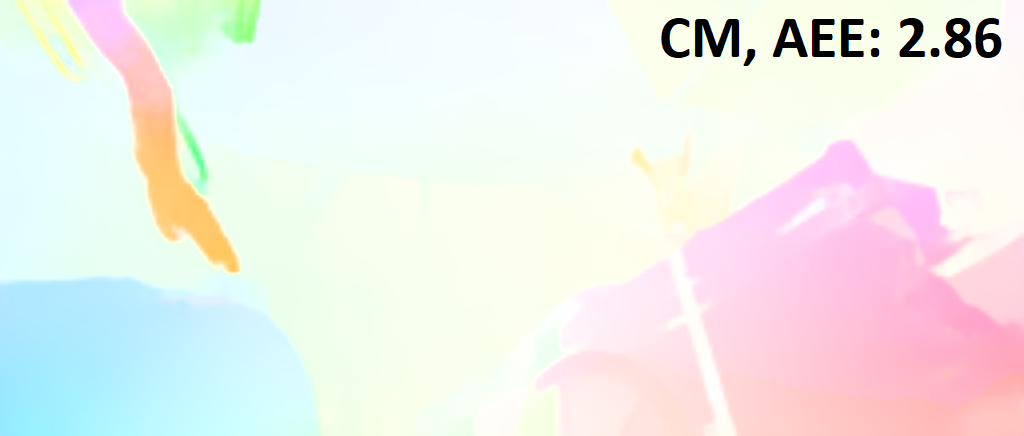}\hfill
   \includegraphics[width=3.05cm]{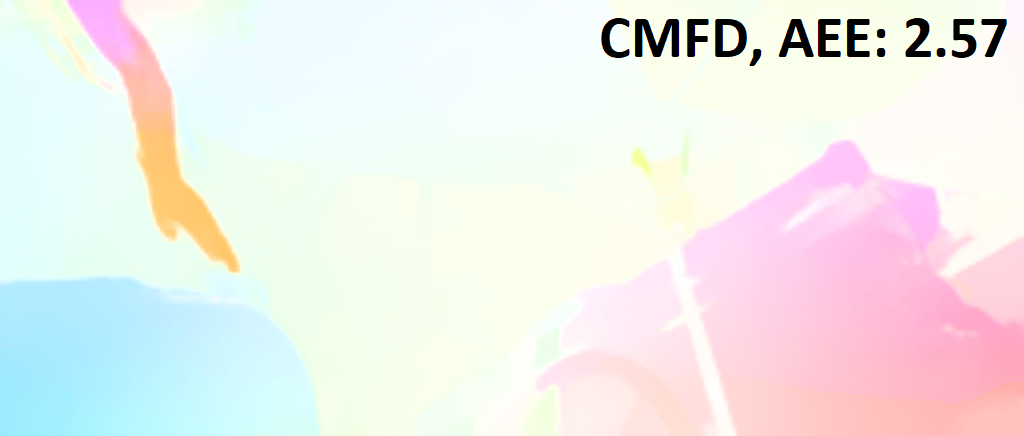}\hfill
   \includegraphics[width=3.05cm]{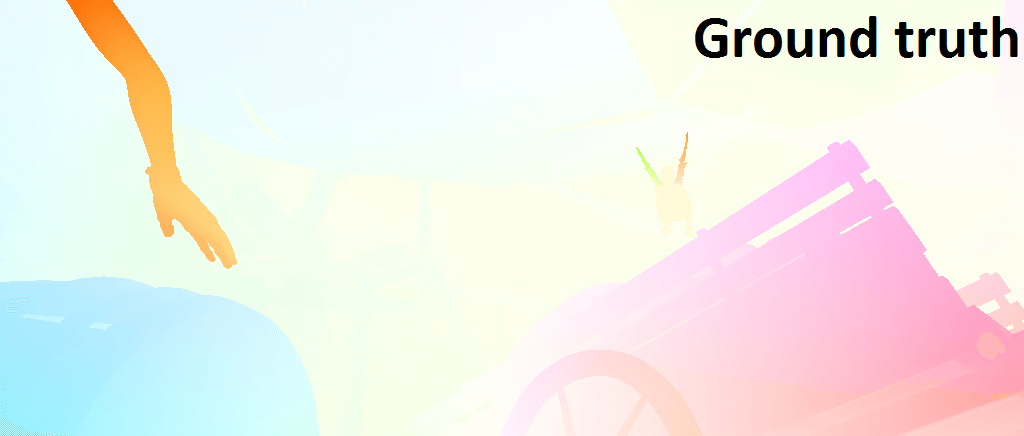}\\

   \includegraphics[width=3.05cm,trim={4.5cm 1cm 4.2cm 0},clip]{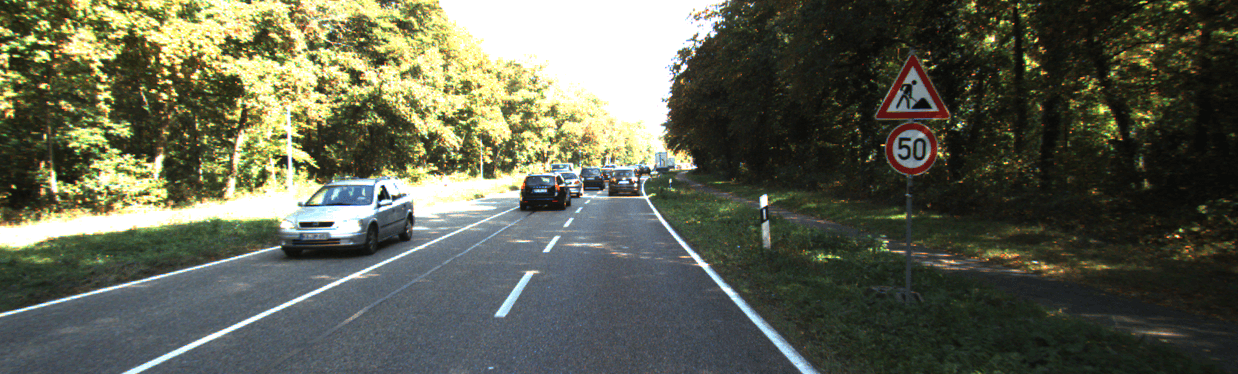}\hfill
   \includegraphics[width=3.05cm,trim={4.5cm 1cm 4.2cm 0},clip]{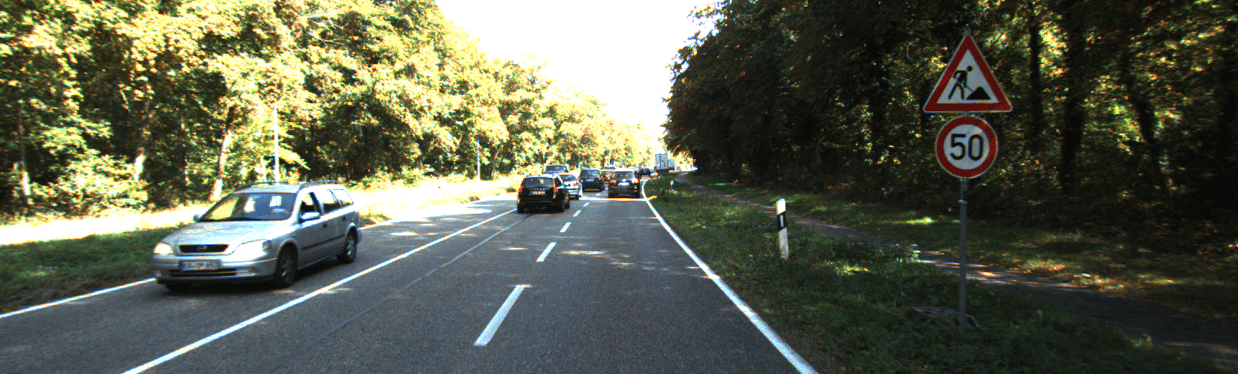}\hfill
   \includegraphics[width=3.05cm,trim={4.5cm 1cm 4.2cm 0},clip]{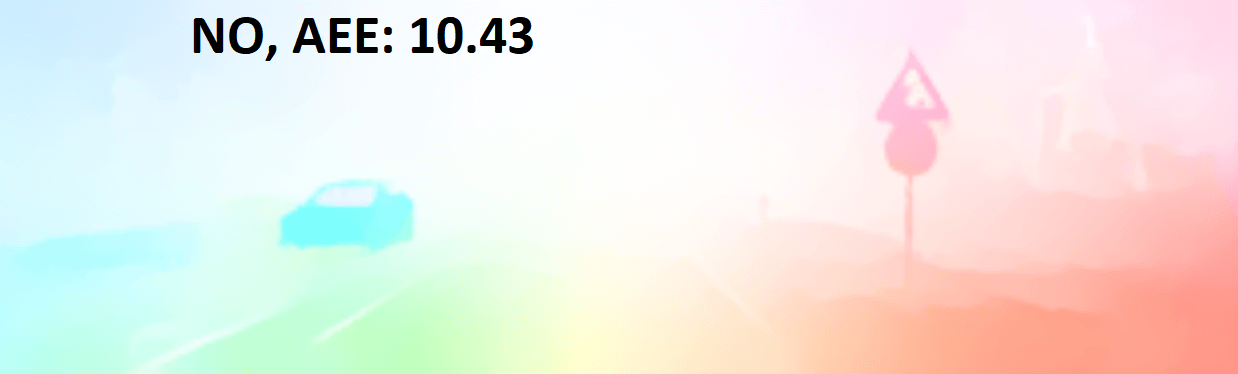}\hfill
   \includegraphics[width=3.05cm,trim={4.5cm 1cm 4.2cm 0},clip]{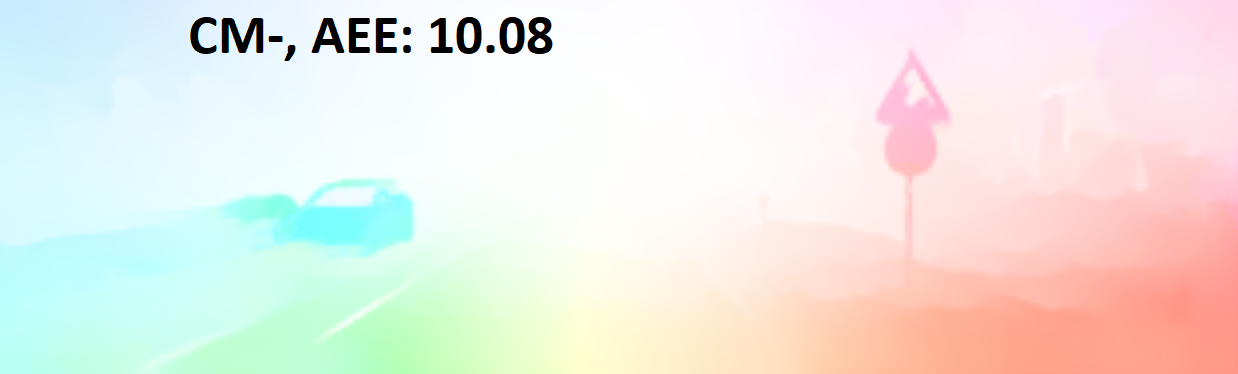}\vspace{-0.1cm}\\

   \includegraphics[width=3.05cm,trim={4.5cm 1cm 4.2cm 0},clip]{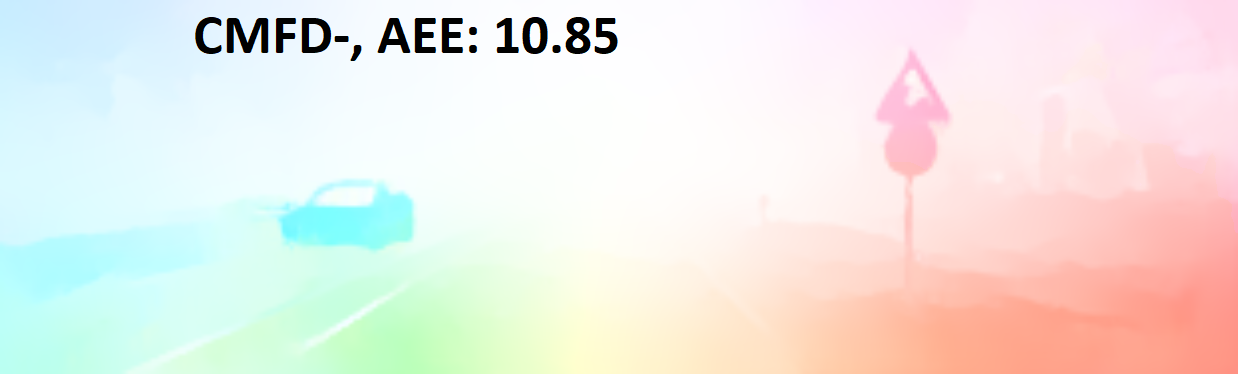}\hfill
   \includegraphics[width=3.05cm,trim={4.5cm 1cm 4.2cm 0},clip]{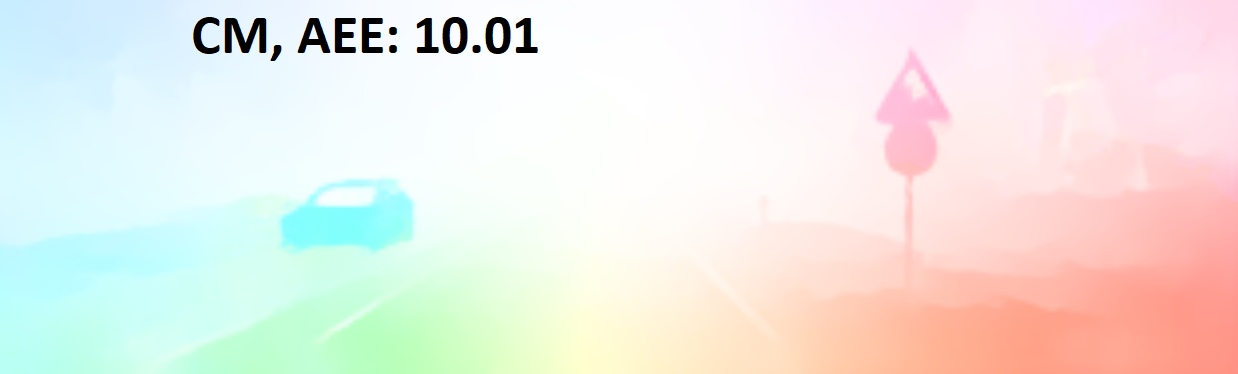}\hfill
   \includegraphics[width=3.05cm,trim={4.5cm 1cm 4.2cm 0},clip]{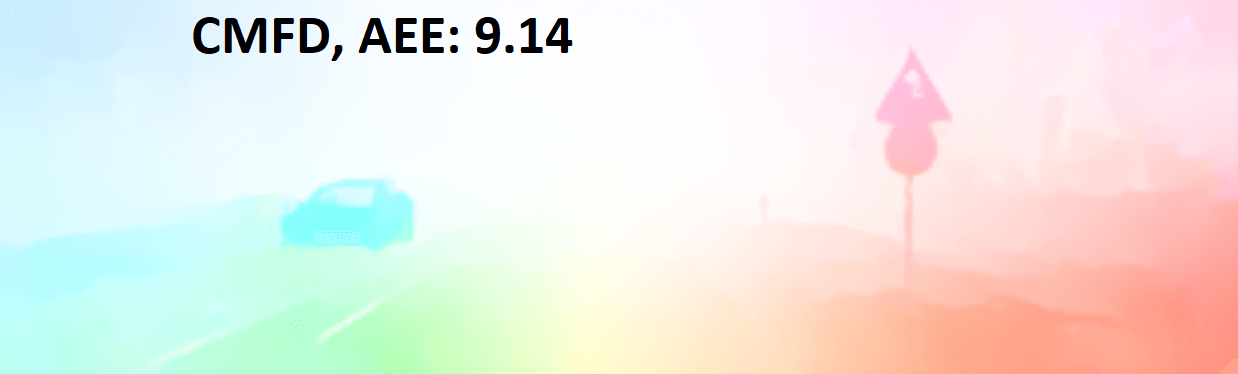}\hfill
   \includegraphics[width=3.05cm,trim={4.5cm 1cm 4.2cm 0},clip]{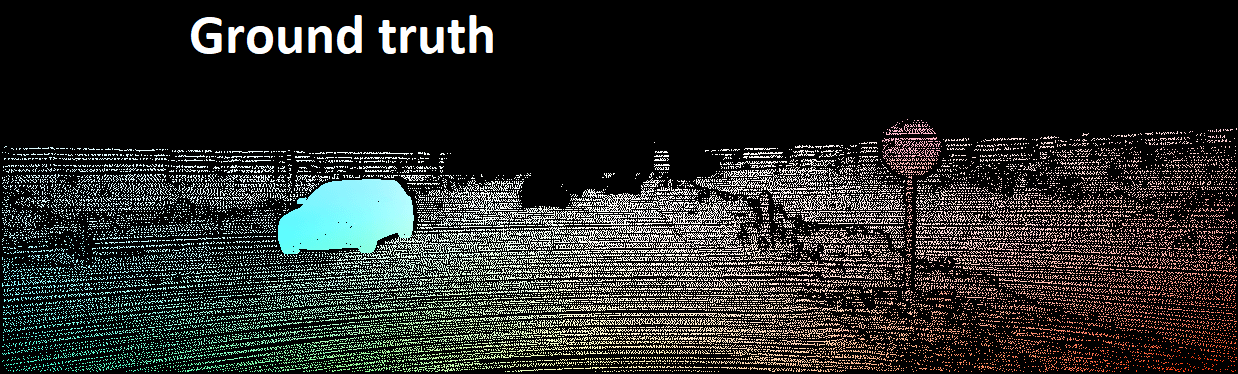}
\end{tabular}
\caption{Examples of flow fields on Sintel Final (top two rows) and KITTI 2015 (bottom two rows) generated by different variants of LiteFlowNet3. Note: NO = No proposed modules are used, CM = \textbf{C}ost Volume \textbf{M}odulation, CMFD = \textbf{C}ost Volume \textbf{M}odulation and \textbf{F}low Field \textbf{D}eformation, and the suffix ``-" indicates that confidence map is not being used.}
\label{fig:ablation study}
\end{figure*}

\begin{figure*}[t]
\captionsetup[subfigure]{labelformat=empty}
\captionsetup[subfloat]{farskip=0pt,captionskip=0pt,justification=centering}
\centering
\begin{tabular}{cccc}
    \subfloat[(a) Confidence map\label{fig:confidence map}]{\includegraphics[width=3.05cm]{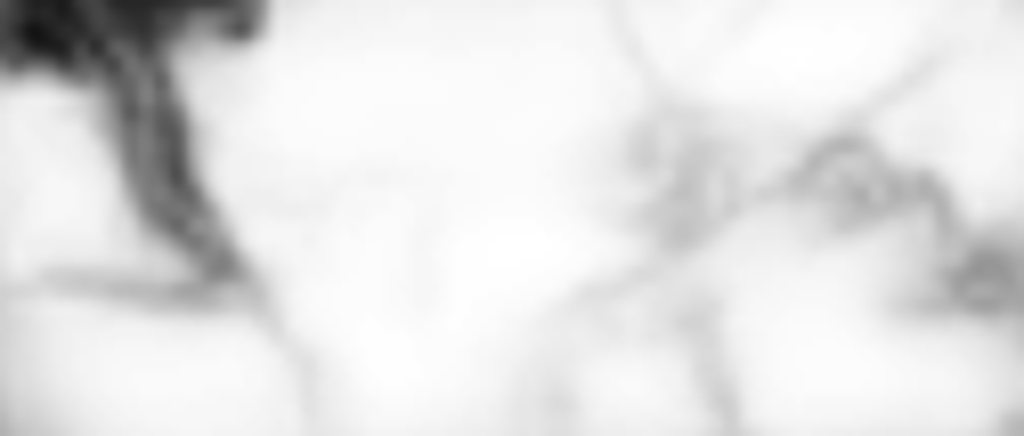}}\hfill
    \subfloat[(b) Original flow field\label{fig:original flow field}]{\includegraphics[width=3.05cm]{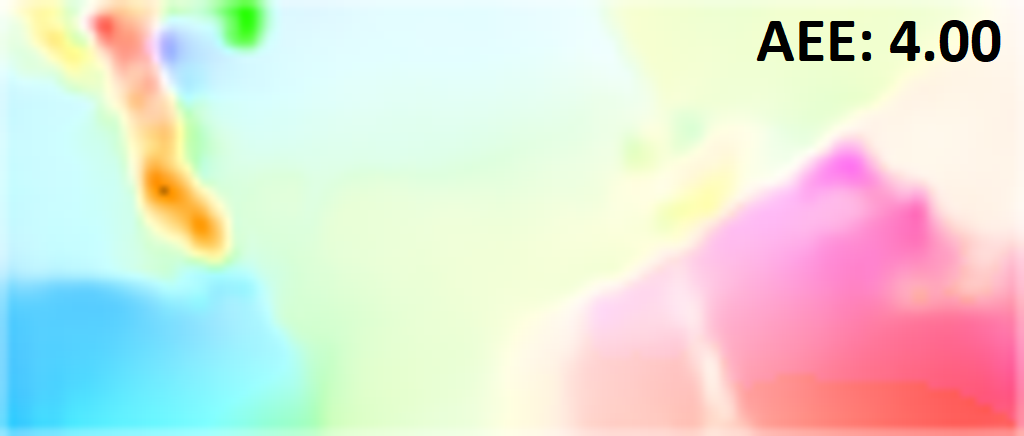}}\hfill
    \subfloat[(c) Displacement\label{fig:displacement field}]{\includegraphics[width=3.05cm]{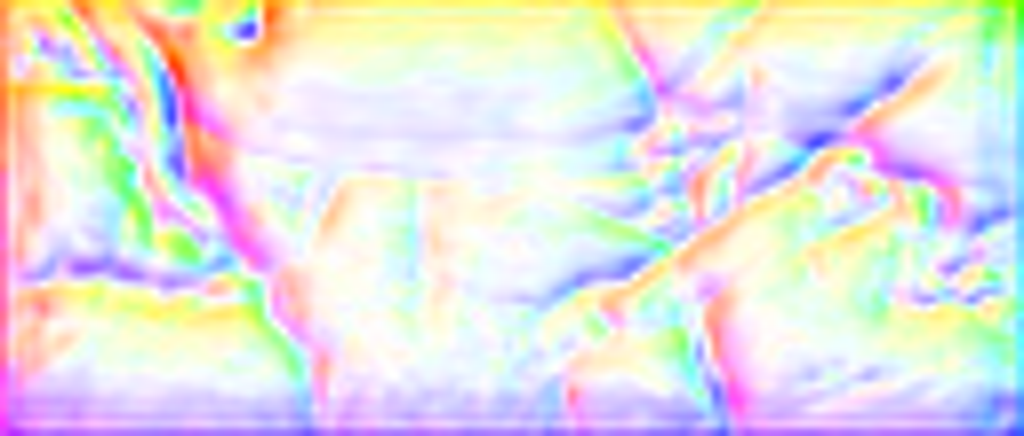}}\hfill
    \subfloat[(d) Deformed flow\label{fig:deformed flow}]{\includegraphics[width=3.05cm]{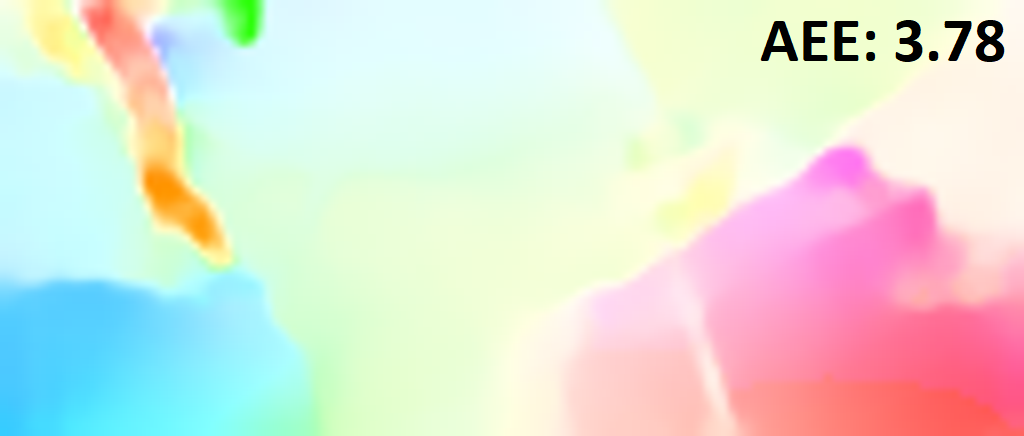}}
\end{tabular}
\caption{An example of flow field deformation. The darker a pixel in the confidence map, the more chance the associated optical flow is not correct.}
\end{figure*}

To study the role of each proposed component in LiteFlowNet3, we disable some of the components and train the resulting variants on FlyingChairs. The evaluation results on the public benchmarks are summarized in Table~\ref{tab:ablation study} and examples of flow fields are illustrated in Fig.~\ref{fig:ablation study}.

\noindent\textbf{Cost Volume Modulation and Flow Deformation.} As revealed in Table~\ref{tab:ablation study}, when only cost volume modulation (CM) is incorporated to LiteFlowNet3, it performs better than its counterpart (NO) neither using modulation nor deformation on all the benchmarks, especially KITTI 2015. When both of cost volume modulation and flow field formation (CMFD) are utilized, it outperforms the others and achieves in a large improvement on KITTI 2015. 
Examples of visual performance are demonstrated in Fig.~\ref{fig:ablation study}. For Sintel, we can observe a large discrepancy in flow color of the human arm between NO and ground truth. On the contrary, flow color is close to ground truth when CM and CMFD are enabled. Particularly, the green artifact is successfully removed in CMFD. 
In the example of KITTI, the car's windshield and triangle road sign in NO are not completely filled with correct optical flow. In comparison with CM, the missed flow can be recovered using CMFD only. This indicates that flow field deformation is more efficient in ``hole filling'' than cost volume modulation.  

\noindent\textbf{Confidence Map.} Variants CM and CMFD, as revealed in Table~\ref{tab:ablation study}, perform better than their counterparts CM- and CMFD- with confidence map disabled.
For the example of Sintel in Fig.~\ref{fig:ablation study}, the green artifact is greatly reduced when comparing CM- with CM. 
Optical flow of the human arm is partially disappeared in CMFD-, while it is recovered in CMFD. The corresponding confidence map is illustrated in Fig.~\ref{fig:confidence map}. It indicates that optical flow near the human arm is highly unreliable.
Similar phenomenon can also be observed in the example of KITTI.
Through pinpointing the flow correctness, the use of confidence map facilitates both cost volume modulation and flow field deformation.

\noindent\textbf{Displacement Field.} As shown in Fig.~\ref{fig:displacement field}, the active region of the displacement field (having strong color intensity) is well-coincided with the active region of the confidence map (having strong darkness, so indicating high probability of being incorrect flow) in Fig.~\ref{fig:confidence map}. 
The deformed flow field in Fig.~\ref{fig:deformed flow} has not only less artifacts but also sharper motion boundaries and a lower AEE when comparing to the flow field without meta-warping in Fig.~\ref{fig:original flow field}.

\section{Conclusion}
Correspondence ambiguity is a common problem in optical flow estimation. Ambiguous feature matching causes outliers to exist in a cost volume and in turn affects the decoding of flow from it. Besides, erroneous optical flow can be propagated to subsequent pyramid levels.
We propose to amend the cost volume prior to the flow decoding. This is accomplished by modulating each cost vector through an adaptive affine transformation. 
We further improve the flow accuracy by replacing each inaccurate optical flow with an accurate one from a nearby position through a meta-warping governed by a displacement field. 
We also propose to use a confidence map to facilitate the generation of modulation parameters and displacement field.
LiteFlowNet3, which incorporates the cost volume modulation and flow field deformation, not only demonstrates promising performance on public benchmarks but also has a small model size and a fast runtime.

\clearpage
%
%
\bibliographystyle{splncs04}
\bibliography{egbib}
\end{document}